\documentclass{article}

    \usepackage[final]{neurips_2025}

\usepackage[utf8]{inputenc} 
\usepackage[T1]{fontenc}    
\usepackage{booktabs}       
\usepackage{microtype}      
\usepackage{xcolor}         
\usepackage{mathtools}
\usepackage{amsfonts}       
\usepackage{nicefrac}       
\usepackage{hyperref}       
\usepackage{url}            
\usepackage{CJKutf8} 
\usepackage{algorithm}
\usepackage{algpseudocode}
\usepackage{enumerate}
\usepackage{adjustbox}
\usepackage{xspace}
\usepackage{amsthm}
\usepackage{verbatim}

\newtheorem{theorem}{Theorem}[section]
\newtheorem{definition}{Definition}[section]
\newtheorem{lemma}[theorem]{Lemma}
\newtheorem*{lemma*}{Lemma}

\newcommand{\repname}{}  
\newtheorem*{reptheoreminner}{\repname}

\newenvironment{reptheorem}[1]{%
  \renewcommand{\repname}{Theorem~\ref{#1}}%
  \begin{reptheoreminner}%
}{%
  \end{reptheoreminner}%
}
\setcitestyle{square}
\setcitestyle{numbers,square}

\title{Estimating Model Performance Under Covariate Shift Without Labels}

\author{Jakub Białek\thanks{These authors contributed equally to this work} \\
	NannyML NV\\
	  Belgium\\
	\texttt{jakub@nannyml.com} \\
    \And
	Juhani Kivimäki\footnotemark[1]\\
	University of Helsinki\\
	Finland\\
	\texttt{juhani.kivimaki@helsinki.fi} \\
	\And
	Wojtek Kuberski\footnotemark[1] \\
	NannyML NV\\
	Belgium\\
	\texttt{wojtek@nannyml.com} \\
	\And
	Nikolaos Perrakis\footnotemark[1]  \\
	NannyML NV\\
	Belgium\\
	\texttt{nikos@nannyml.com}
}

\begin{document}

\maketitle

\begin{abstract}
After deployment, machine learning models often experience performance degradation due to shifts in data distribution. It is challenging to assess post-deployment performance accurately when labels are missing or delayed. Existing proxy methods, such as data drift detection, fail to measure the effects of these shifts adequately. To address this, we introduce a new method for evaluating binary classification models on unlabeled tabular data that accurately estimates model performance under covariate shift and call it Probabilistic Adaptive Performance Estimation (PAPE). It can be applied to any performance metric defined with elements of the confusion matrix. Crucially, PAPE operates independently of the original model, relying only on its predictions and probability estimates, and does not need any assumptions about the nature of covariate shift, learning directly from data instead. We tested PAPE using over 900 dataset-model combinations from the US census data, assessing its performance against several benchmarks through various metrics. Our findings show that PAPE outperforms other methodologies, making it a superior choice for estimating the performance of binary classification models.
\end{abstract}

\section{Introduction}

The final step in the machine learning (ML) model development process is to evaluate the model on an unseen dataset, known as \textit{test set}. Successful performance on this dataset, indicating potential business value, typically leads to the model’s deployment to \textit{production}. The model’s performance on the production data is expected to match the performance measured on the test set~\cite{lones:2021}. Yet, this assumption often fails due to data distribution shifts that lead to model performance degradation~\cite{vela:2022}. Therefore, ongoing monitoring of the model’s performance is essential to ensure it continues to meet expected outcomes. However, measuring real-time performance is challenging in many scenarios because true labels are either unavailable or significantly delayed.

In scenarios where labels are unavailable, performance is commonly estimated using proxy methods that primarily monitor changes in the input data distribution~\cite{klaise:2020}. However, such changes in input data often have only a minimal effect on the model's actual performance~\cite{rabanser:2019}, and even when there is a notable effect, it might not be harmful~\cite{ginsberg:2022}. Additionally, the methods used to measure changes in the input distribution typically provide estimates in units that do not correspond to performance metrics. Thus, at best, input data monitoring can offer some qualitative understanding of the performance stability. When a distribution shift does occur, these methods do not provide any insight into the direction or magnitude of the impact on performance. The true extent of performance changes only becomes apparent once the labels become available, which may sometimes never happen. Until then, the model might incur significant financial losses~\cite{tan:2022}.

In recent years, a new approach of \textit{unsupervised accuracy estimation}~\cite{chen:2021} has produced several methods to estimate model performance directly without access to labels. A notable shortcoming of these methods has been that they focus solely on estimating the accuracy of a given classifier model. However, accuracy is often not the most appropriate metric for assessing model performance, which has recently motivated the shift of focus to \textit{unsupervised performance estimation}~\cite{kivimaki:2025b}, where estimators should be applicable to any classification metric, not just accuracy. In this paper, we extend this line of work and present
Probabilistic Adaptive Performance Estimator (PAPE), a novel method for 
estimating any classification metric that can be defined using the elements of the \textit{confusion matrix}. PAPE is provably robust against \textit{covariate shift}. Furthermore, PAPE models the full distribution of the estimated metric, which can be leveraged to produce valid confidence intervals for the estimates.

With these estimates, the impact of changes in model performance on business operations can be quantified, allowing informed decision-making about corrective adaptations. These may include altering how model predictions are utilized in downstream processes, determining if model retraining is required, or whether some fail-safe mechanism should trigger.

The main contributions of this paper are the following:
\begin{itemize}
\item We introduce PAPE, a novel method for estimating model performance under covariate shift. PAPE can be applied to any classification metric that can be defined with elements of the confusion matrix. Importantly, PAPE autonomously learns from the data without any user input regarding the nature of the covariate shift.
\item We provide theoretical guarantees for the estimation quality of PAPE and derive bounds for its estimation error under approximate calibration for composable metrics.
\item We demonstrate PAPE's effectiveness empirically. Our experiments show that PAPE significantly outperforms all previous methods across all metrics evaluated.

\end{itemize}


\section{Related Work}\label{sec:work}

In recent years, there has been a surge of new suggested methods seeking to solve the unsupervised accuracy estimation problem. Importance Weighting (IW)~\cite{shimodaira:2000} is a simple but powerful method used in model adaptation~\cite{lu:2022} that can also be used to estimate model performance. IW calculates a likelihood ratio of observing test set input data in production and uses this ratio to estimate performance for production data as a weighted metric calculated for the test set data. An extended version of IW~\cite{chen:2021b} incorporates knowledge about the differences between test and production data distributions and their impact on performance, but requires the user to specify this information a priori. 

A set of methods is based on training an ensemble of models and comparing their predictions~\cite{baek:2022,chen:2021,jiang:2022}. Other approaches requiring additional training include Contrastive Automated Model Evaluation (CAME)~\cite{peng:2023}, where the model training objective is augmented with a contrastive learning component, and reverse testing (RT)~\cite{fan:2006}, which uses the monitored model predictions as labels to train a reverse model on production data. The reverse model’s performance on the test dataset is then assumed to indicate the monitored model’s performance on the production data. Whilst these approaches offer promising results, they cannot be applied off the shelf but require additions or alterations in training the model being estimated, and often come with a significant computational overhead.

Another set of methods measures distributional distances between the test and production data distributions~\cite{guillory:2021,deng:2021,deng:2023,lu:2023} and either tries to identify the change in performance directly based on this distance or by training a light-weight regressor model that is used to map the distance into a change in performance. The main challenge of these methods is in translating the measured distance into a meaningful performance estimate, either requiring the creation of multiple synthetic shifted data distributions to train the regressor~\cite{guillory:2021} or resorting to quantifying the shift in performance in terms of some secondary metric such as correlation~\cite{deng:2023}.

In this work, our main interest lies in methods that utilize the model's confidence in estimating its performance. Average Confidence (AC)~\cite{hendrycks:2017} was originally suggested for Out-of-Distribution~(OoD) detection but has since become the de facto baseline of confidence-based estimators~\cite{kivimaki:2025}. Difference of Confidence (DoC)~\cite{guillory:2021} uses the difference between confidence scores from test to production data and fits a regression model that maps this difference to accuracy. Average Thresholded Confidence (ATC)~\cite{garg:2022} learns a threshold on confidence scores and estimates accuracy as the fraction of predictions that exceed this threshold. Confidence Optimal Transport (COT)~\cite{lu:2023}
measures prediction error as a Wasserstein distance between the test set's label distribution and the production set's pseudo-label distribution. 

A key limitation of all of the above methods is that they are developed to estimate only classification accuracy, whereas other classification metrics are often more appropriate and descriptive of model performance. For example, in situations with severe class imbalance, a dummy model, which always predicts the majority class, can achieve deceptively high accuracy~\cite{bekkar:2013}. Recently, a new approach of Confidence-based Performance Estimation (CBPE)~\cite{kivimaki:2025b} has been proposed to address this shortcoming of previous estimators. 

Since PAPE is essentially an extension of the CBPE method, we describe its key properties here (see also Appendix~\ref{app:implementation}). In CBPE, confidence scores produced by the model for a sample of predictions are used to estimate the elements of the confusion matrix. Each element is treated as a random variable following a Poisson binomial distribution whose full probability distribution is derived using the confidence scores as parameters. Using the expected values of the distributions as point estimates for the elements of the confusion matrix, these estimates have been shown to be unbiased and consistent under perfect calibration~\cite{kivimaki:2025b}. Having estimated the elements of the confusion matrix, one can take any classification metric that can be defined in terms of these elements and derive the distribution of the metric based on the distributions of the elements using an appropriate algorithm~\cite{kivimaki:2025b}. Again, the expected value of this distribution can then be used as a point estimate for that metric. Additionally, one can produce valid confidence intervals for these metrics from the derived probability distributions~\cite{kivimaki:2025b}. 

Although CBPE is a theoretically sound approach to solving the unsupervised performance estimation problem, its main downside is its reliance on model calibration, which is known to deteriorate under certain distributional shifts~\cite{ovadia:2019}. This phenomenon undermines the usability of CBPE and all other confidence-based estimators in a way that is currently not well understood. There have even been conflicting reports on whether calibration in confidence-based estimators is useful or not~\cite{guillory:2021,garg:2022}. In this work, we offer a remedy to this problem by augmenting the CBPE approach to make its calibration more robust against distributional shifts, resulting in PAPE, which we will describe in the following section.

\section{Methodology}\label{sec:methodology}

In this section, we present PAPE, a new algorithm for estimating model performance under covariate shift. We begin by describing the setting in Section~\ref{sec:setting}. Then, in Section~\ref{sec:calibration}, we define $\alpha$-approximate calibration along with some theoretical insights. Finally, In Section~\ref{sec:PAPE}, we describe the PAPE method and give a bound for its estimation error for certain metrics.

\subsection{Unsupervised Performance Estimation Under Covariate Shift}\label{sec:setting}

Suppose we have trained a probabilistic binary classifier, where $f: \mathcal{X} \rightarrow [0,1]$ outputs a confidence score for the positive class and $g: [0, 1] \rightarrow \{0, 1\}$ maps these scores to binary predictions. The classifier is trained using data from some source distribution $\mathcal{D}_s=p_s(\boldsymbol{x}, y)$, where we have access to labels. We further assume that the image of $f$ is a countable set $R(f)$, where each value $v\in R(f)$ defines a \textit{levelset} $\{\boldsymbol{x} \in \mathcal{X}: f(\boldsymbol{x})=v\}$. We seek to estimate our model's performance in a potentially shifted target distribution $\mathcal{D}_t=p_t(\boldsymbol{x}, y)$, where we have access only to samples from $p_t(\boldsymbol{x})$. 

If no assumptions about the nature of the shift are made, the unsupervised performance estimation task is impossible~\cite{david:2010,lu:2023}. In this work, we make the most commonly used \textit{covariate shift assumption}~\cite{moreno-torres:2012}, where the shift from the source to target distribution can be attributed solely to changes in the marginal distribution of the covariates. That is, since any distribution can be factorized as $p(\boldsymbol{x}, y) = p(y|\boldsymbol{x})p(\boldsymbol{x})$, we assume that the label-assigning conditional distribution remains unchanged $p_s(y|\boldsymbol{x}) = p_t(y|\boldsymbol{x})$ while the marginal distribution of covariates shifts $p_s(\boldsymbol{x}) \neq p_t(\boldsymbol{x})$.

\subsection{Approximate Confidence Calibration}\label{sec:calibration}

As mentioned, a key shortcoming with confidence-based estimators has been the dependence on a small calibration error, which $f$ is not guaranteed to achieve out-of-the-box. The simplest way to fix this is to use a small amount of validation data from the source distribution to train a post-hoc calibration mapping $c:[0,1] \rightarrow [0,1]$ to minimize the expected absolute calibration error $\mathbb{E}_{p_s(\boldsymbol{x})}\left[|P_{\mathcal{D}_s}(Y=1 \mid c(f(X))) - c(f(X))|\right]$. Unfortunately, even if calibration error is eliminated in the source distribution, it typically remanifests in the target distribution under covariate shift~\cite{ovadia:2019}. 

Previous research has provided theoretical guarantees for confidence-based performance estimation in the ideal situation of perfect calibration~\cite{kivimaki:2025b}. In this work, we generalize this scope to approximate calibration. Recently, in AI-fairness research, researchers have borrowed ideas from the field of differential privacy~\cite{hebert:2018}, which has led to the notion of $\alpha$-approximate calibration defined as follows:
\begin{definition}
    Assume $f$ is a model $f: \mathcal{X} \rightarrow [0, 1]$ with a countable set of values $R(f)$, and that $\alpha \ge 0$. We say that $f$ is $\alpha$-approximately calibrated in $\mathcal{D}$, if:
    \begin{equation}\label{def:calibration}
        K(f, \mathcal{D})=\sum_{v \in R(f)}P_{\mathcal{D}}(f(\boldsymbol{x})=v) \bigg|\mathbb{E}_{\mathcal{D}}[y  \mid f(\boldsymbol{x})=v] - v \bigg| \le \alpha.
    \end{equation}
\end{definition}
Note that by setting $\alpha=0$, we get the commonly used definition of perfect calibration~\cite{guo:2017} (marginalized over all possible values $f$ can take) as a special case. Calibration in this sense is a marginal guarantee, since the left-hand side of the Inequality~(\ref{def:calibration}) is an average over the whole distribution. In fairness research, it has become apparent that such guarantees should also apply conditionally on some properties of the instance, such as group membership. Otherwise, a model might be well-calibrated overall, but yield a high bias for some (possibly protected) minority groups. This has led to the notion of \textit{multicalibration}, which we will define in its general form.
\begin{definition}
    Assume $f$ is a model $f: \mathcal{X} \rightarrow [0, 1]$ with a countable set of values $R(f)$, $\mathcal{H}$ is a collection of functions $h: \mathcal{X} \rightarrow \mathbb{R}$, and $\alpha \ge 0$. We say that $f$ is $\alpha$-approximately multicalibrated in $\mathcal{D}$ with respect to $\mathcal{H}$, if $~\forall h \in \mathcal{H}$:
    \begin{equation}
        K(f, \mathcal{D}, \mathcal{H})=\sum_{v \in R(f)}P_{\mathcal{D}}(f(\boldsymbol{x})=v) \bigg|\mathbb{E}_{\mathcal{D}}[h(\boldsymbol{x})(y - v) \mid f(\boldsymbol{x})=v]\bigg| \le \alpha.
    \end{equation}
\end{definition}

Here, the functions $h$ were originally indicator functions for group membership for groups of interest, but this was later generalized to any real-valued functions to allow for weighted memberships~\cite{globus:2023}. Recently, $\mathcal{H}$ was used as a hypothesis space for \textit{density ratio estimation} (DRE) models, which are used to approximate the true likelihood ratios $w_{s \rightarrow t}(\boldsymbol{x}) = \frac{p_t(\boldsymbol{x})}{p_s(\boldsymbol{x})}$ under subpopulation shift~\cite{kim:2022}. For this setting, we have the following theorem:
\begin{theorem}\label{thr:calibration}
    Assume that $p_s(y|\boldsymbol{x}) = p_t(y|\boldsymbol{x})$ and that $f$ is $\alpha$-approximately multicalibrated in $\mathcal{D}_s$ with respect to $\mathcal{H}$. If $w_{s \rightarrow t} \in \mathcal{H}$, then $K(f, \mathcal{D}_t) \le \alpha$.
\end{theorem}
In addition to providing a proof of this theorem in Appendix~\ref{app:proofs:th3_1}, we will also give an upper bound for the calibration error in $\mathcal{D}_t$ when $w_{s \rightarrow t} \notin \mathcal{H}$ and some $h \in \mathcal{H}$ is used to approximate $w_{s \rightarrow t}$ instead (in Appendix~\ref{app:proofs:imperfect_weights}). Multicalibration in this setting is used to anticipate potential changes in the marginal distribution $p(\boldsymbol{x})$ a priori and to approximately adapt to all such changes simultaneously. Although algorithms for achieving multicalibration exist, they are computationally demanding~\cite{hebert:2018}. This problem becomes increasingly difficult with the size of $\mathcal{H}$~\cite{globus:2023}, which can be infinite when considering all possible ways a distribution can shift. PAPE circumvents this problem by extracting data from an actual shifted distribution and fitting a calibration mapping to exactly that distribution, essentially becoming multicalibrated with respect to an $\mathcal{H}$ that has only one member. We will describe the details of this process next.

\subsection{Probabilistic Adaptive Performance Estimation (PAPE)}\label{sec:PAPE}
In this section, we introduce PAPE and explain how it extends CBPE by addressing CBPE’s main limitation: maintaining calibration under covariate shift. Let $\hat{Y}=g(f(X))$ denote the binary prediction and $m: \mathcal{Y} \times \mathcal{Y} \rightarrow \mathbb{R}$ be a binary classification performance metric with some unknown expected value $m_{(f,g,\mathcal{D}_t)}=\mathbb{E}_{\mathcal{D}_t}[m(\hat{Y}, Y)]$ in distribution $\mathcal{D}_t$, where we don't have access to labels. However, we do have access to labels in some source distribution $\mathcal{D}_s$. Assume that $f$ is already trained with data from $\mathcal{D}_s$. We start by collecting (i.i.d.) random samples $\{(X_i^s, Y_i^s)\}_{i=1}^{n_s} \sim p_s(\boldsymbol{x}, y)$ and $\{X_j^t\}_{j=1}^{n_t} \sim p_t(\boldsymbol{x})$, and training a DRE model from a hypothesis space of binary classifiers $\mathcal{H} \subseteq \{h\mid h:\mathcal{X}\rightarrow \{0,1\}\}$ defined by the learning algorithm of user's choice as follows. 

First, we assign indicator labels $z_i^{s}=0$ to all $X_{i}^s$ and $z_j^{t}=1$ to all $X_{j}^t$. 
Next, we concatenate the features $X^{st}=[X^s; X^t]\in \mathbb{R}^{(n_s+n_t)\times d}$ (where $d$ is the dimensionality of $\mathcal{X}$) and their corresponding indicator labels $z^{st}=[z^s; z^t]\in \mathbb{R}^{n_s+n_t}$.
Then, we use $(X^{st}, z^{st})$ as training data for the DRE model that learns to discriminate between samples from the source and target distributions as described in~\cite[chapter 2.7.5]{murphy:2023}. Once the best-fit DRE model $h^* \in \mathcal{H}$ is found, we can use it to estimate the probabilities of observing instances $\boldsymbol{x}_i^s$ from the source distribution within the target distribution, formally $P(z_i=1 \mid X_i^s=\boldsymbol{x}_i^s) \approx h^*(\boldsymbol{x}_i^s)$. Finally, we can approximate $w_{s \rightarrow t}$ with
\begin{equation}
    \widehat{w}_{s \rightarrow t}(\boldsymbol{x}_i^s) = \frac{n_s}{n_t}\cdot\frac{h^*(\boldsymbol{x}_i^s)}{1-h^*(\boldsymbol{x}_i^s)},
\end{equation}
and train a weighted calibration mapping $c$ by fitting it to $\{(f(X_i^s), Y_i^s)\}_{i=1}^{n_s}$ using $\widehat{w}_{s \rightarrow t}(\boldsymbol{x}_i^s)$ as weights.
Once the calibrated scores $c(f(X_j^t))$ are available, they can be used to derive performance estimates with CBPE without access to labels from the target distribution~\cite{kivimaki:2025b}. 

As a side note, one would typically choose a dedicated calibration mapping~\cite{platt:1999, zadrozny:2001, kull:2017} for this purpose, as they enforce monotonicity, keeping the ranking of the scores and the resulting binary predictions unchanged. However, with PAPE, one can use any regression model of choice as the calibration mapping since the calibrated scores are used solely for performance estimation purposes and don't affect the binary predictions in any way. 

If $c \circ f$ is perfectly calibrated in $\mathcal{D}_t$, the theoretical guarantees from CBPE~\cite{kivimaki:2025b} carry over, and PAPE estimates are guaranteed to be unbiased and consistent. However, since perfect calibration is unattainable in any real-life situation, let us next explore the relation between calibration error and PAPE estimation error in a limited setting: Some performance metrics, such as accuracy and precision, can be calculated as a mean of observation-level values. We call such metrics \textit{composable}. For any composable metric $m$, the PAPE estimate can be written as
\begin{equation}
	\widehat{m}_{(c \circ f, g, \mathcal{D}_t)} = 
	\mathbb{E}_{p_t(\boldsymbol{x})}
	\left[ 
	c(f(\boldsymbol{x}))m\left(\hat{y},1\right) +
	\left(1 - c(f(\boldsymbol{x}))\right)m\left(\hat{y},0\right)
	\right]
	\label{eq:MCBPE}.
\end{equation}
In practice, this expectation is approximated with the sample mean. For any composable metric, the estimation error of PAPE is bounded by the calibration error as described by the following theorem, which we will prove in Appendix~\ref{app:proofs:th3_2}:

\begin{theorem}\label{thr:estimation}
Let $c \circ f$ be $\alpha$-calibrated in $\mathcal{D}_t$. Also, assume that $f$ has a countable image set $R(f)$, $m$ is a composable metric with $0 \le m(\hat{y}, y) \le 1$, and that $\hat{m}$ is its PAPE estimate. Then,
    \begin{equation*}
    |m_{\left(f, g, \mathcal{D}_t\right)} - \widehat{m}_{(c \circ f, g, \mathcal{D}_t)}| \le \alpha.
    \end{equation*}
\end{theorem}

For metrics that are not composable, such as the F$_1$ score, the above bound is not guaranteed to hold. However, our empirical experiments show that the PAPE estimates are superior to any other unsupervised performance estimation method for these metrics as well. We will present these findings next.

\section{Experimental Evaluation}\label{sec:experiments}

This section describes our experimental setting, where we evaluated and compared the proposed method against existing benchmarks. We describe the datasets we used in Section~\ref{sec:datasets}. The experimental setup, along with the ML models whose performance was estimated, is described in Section~\ref{sec:expsetup}. We provide practical implementation details of the evaluated benchmarks in Section~\ref{sec:benchmarks} (see also Appendix~\ref{app:implementation}). In Section~\ref{sec:evaluation} we propose novel evaluation metrics suitable for aggregating over multiple dissimilar evaluation cases and present the results of our experiments in terms of these metrics. We ran additional experiments with data from the recently published TableShift benchmark~\cite{gardner:2024} and explored the effect of sample size on the estimators. These experiments are described in Appendix~\ref{app:experiments}.

\subsection{Datasets}\label{sec:datasets}

The datasets we used to evaluate the method come from Folktables~\cite{folktables}. Folktables uses US census data and preprocesses it to create a set of binary classification problems. We used the following tasks: ACSIncome, ACSPublicCoverage, ACSMobility, ACSEmployment, ACSTravelTime. For each of the five prediction tasks listed above, we fetched Folktables data for all 50 US states
spanning four consecutive years (2015-2018). This gave us 250 datasets.

\subsection{Experimental Setup}\label{sec:expsetup}

We started by separating the first-year data (2015) in each fetched dataset and used it as a training period. For each resulting training data set, we fitted five commonly used binary classification algorithms: Logistic Regression, Neural Network Model~\cite{gorishniy:2021}, Random Forest~\cite{scikit-learn}, XGBoost~\cite{xgboost}, and LGBM~\cite{ke2017lightgbm} with default parameters. We used these models to predict labels on the remaining part of the datasets. We recorded both binary predictions and confidence scores for the positive class. This resulted in 1,250 dataset-model pairs, which we call evaluation cases.

The rest of the data for each case was further divided into two periods - \textit{reference} (the year 2016) and \textit{production} (2017, 2018). Reference data was used to fit the performance estimation methods with model inputs, model outputs, and actual labels. Production data was further split into data chunks of 2,000 observations each, maintaining the order of the observations. The realized performance of the monitored model was recorded for each data chunk based on the monitored model's outputs and actual labels. The performance of the monitored model for the same chunks was then estimated based on the monitored model's inputs and outputs. 

\begin{figure}[ht!]
	\centering
	\includegraphics[width=1.0\textwidth]{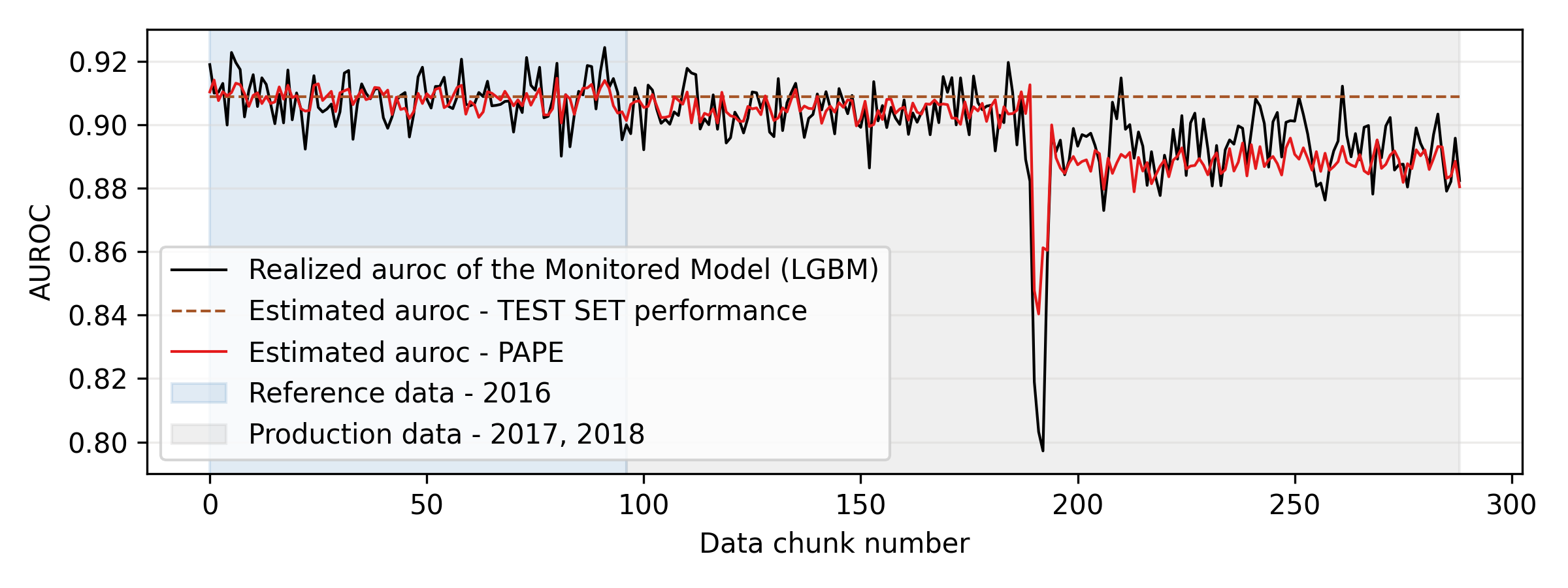}
	\caption{
		 Estimation of AUROC for ACSIncome data (California) and LGBM as the monitored model. The black line is the realized AUROC of the monitored model for each data chunk. The red line is the AUROC estimated with PAPE. The brown dashed line is the TEST SET performance.
	}
	\label{fig:expsetup}
\end{figure}

For each production data chunk, we compared the realized performance and the estimated performance for each performance estimation method. We filtered out evaluation cases with fewer than 6,000 observations (3 chunks) in the reference data set and cases where the performance of the monitored model on the reference data was worse than random (that is, with an AUROC lower than 0.5). After this filtering, we ended up with 959 evaluation cases, containing 36,557 production data chunks (evaluation points) for each evaluated method\footnote{We used a single 11th Gen Intel i7-11800H 2.30GHz machine; computation took over 120 hours.}. We measured the estimation performance of each method with three metrics: Accuracy, F$_1$ score, and AUROC. Figure \ref{fig:expsetup} shows an example of an AUROC estimation result.

\subsection{Benchmarks}\label{sec:benchmarks}

Below, we describe the benchmarks that were evaluated against PAPE and their implementation details. 
We omit MANDOLINE~\cite{chen:2021b} as it requires user input on the nature of the covariate shift.

\paragraph{PAPE} We implement PAPE as described in Section~\ref{sec:PAPE}. We use an LGBM~\cite{ke2017lightgbm} Classifier as the DRE model, and an LGBM Regressor for the calibration mapping, both with default hyperparameters. We assume that the reference data originates from the source distribution $\mathcal{D}_s$ and that each production data chunk originates from some target distribution $\mathcal{D}_t$, which is potentially different for each chunk. 

\paragraph{TEST SET performance} For each evaluation data chunk from production data, the performance estimated by this benchmark equals the performance calculated on reference data (and typically, the test set is chosen as the reference set). It is our baseline benchmark, representing the initial assumption under which the model's performance on the production data is constant and equal to the performance calculated on the test set with reference data.

\paragraph{Confidence-based Performance Estimation (CBPE)}
CBPE~\cite{kivimaki:2025b} is a simpler version of PAPE. We train a calibration mapping as with PAPE, using reference data from $\mathcal{D}_s$, but we do not use likelihood ratios in training the mapping (it is equivalent to PAPE with $w_{s \rightarrow t}$ fixed to $1$ for all observations). This results in a classifier that has a small calibration error with the reference data, but is not guaranteed to be well-calibrated with the production data from $\mathcal{D}_t$.

\paragraph{Average Threshold Confidence (ATC)} With ATC~\cite{garg:2022}, we take the raw scores provided by the monitored model \(f(\boldsymbol{x})\) and apply the maximum confidence function to them, denoting it as \(\mathrm{MC}\):
\begin{equation}
	\label{eq:mc}
	\mathrm{MC}\left(f(\boldsymbol{x})\right) = 
	\begin{cases}
		f(\boldsymbol{x}),	& f(\boldsymbol{x}) \geq 0.5\\
		1-f(\boldsymbol{x}),	& \text{otherwise}
	\end{cases}
\end{equation}
Then we use it to learn a threshold on reference data such that the fraction of observations above the threshold is equal to the performance metric value calculated on reference data. When inferring, we apply MC to the evaluation data chunk and calculate the fraction of observations above the learned threshold. The estimated metric is equal to the calculated fraction.  

\paragraph{Difference of Confidence (DoC)} DoC~\cite{guillory:2021} assumes that the performance change is proportional to the change in the mean of maximum confidence \eqref{eq:mc}. To learn this relationship, we fit a Linear Regression model on the difference of confidence between the shifted and the reference data as input, and the difference in performance between these two as the target. In order to get datasets with synthetic distribution shifts, we perform multiple random resamplings of the reference dataset.

\paragraph{Confidence Optimal Transport (COT)}
In binary classification, COT~\cite{lu:2023} finds the optimal transport plan from the scores $f(X_i^t)$ with $\{X_i^t\}\sim p_t(\boldsymbol{x})^n$ to the labels $y_j^s$ with $\{y_j^s\}\sim p_s(y)^n$, where $n$ is the chunk size. The cost of this transportation plan is used as an estimate of the classification error of $f$ in $\mathcal{D}_t$. We estimate only accuracy with this method (see Appendix~\ref{app:implementation} for details).

\paragraph{Modified Reverse Testing (RT-mod)}
Reverse Testing methods train a \textit{reverse} model on production data with the monitored model inputs as features and monitored model predictions as targets. This reverse model is then used to make predictions on reference inputs, and its performance is evaluated with the reference labels. The reverse model’s performance on reference data is a proxy for the monitored model’s performance on the production data. We modify the method by additionally fitting Linear Regression to relate the reverse model performance change with the monitored model performance (just like in DoC).

\paragraph{Importance Weighting (IW)}
For IW, we first estimate density ratios $w_{s \rightarrow t}$ between reference and production data with the DRE classifier, the same as we use for PAPE. Then, we use them as weights to estimate the weighted performance metric of interest using reference data.

\subsection{Evaluation}\label{sec:evaluation}

Performance estimation is a regression problem, which motivates the use of evaluation metrics such as the Mean Absolute Error (MAE) or Root Mean Squared Error (RMSE). However, aggregating MAE/RMSE over multiple models and chunks of data from different datasets leads to skewed and uninterpretable results. Large MAE/RMSE for an evaluation case where the model's performance has a large variance might still be satisfactory. On the other hand, even small changes (in the absolute scale) in performance might be significant in cases where the model's performance is very stable.

To account for this scaling issue, we used normalized versions of MAE and RMSE, where we scaled absolute/squared errors by the standard error (SE) calculated for each evaluation case. We first measured SE as the standard deviation of the realized performance sampling distribution on the reference data. We did this by repeatedly sampling 2,000 observations (the size of the evaluation data chunk) at random from the reference data with replacement. Then, we calculated the realized performance metric on each sample. We repeated this 500 times and calculated the standard deviation of the obtained per-sample metrics - the standard error (SE). Then, for each evaluation case, we divided the MAE and RMSE by SE and squared SE, respectively, resulting in normalized mean absolute error (NMAE) and normalized root mean squared error (NRMSE), defined formally as
\begin{align*}
\mathrm{NMAE} & =
	\frac{1}{\sum^k_{i=1} c_i}
	\sum\limits_{i=1}^k \sum\limits_{j=1}^{c_i} 
	\frac{\left| m_{i,j} - \widehat{m}_{i,j}\right|}{\mathrm{SE}_i}\\
	\mathrm{NRMSE} & = \sqrt{
		\frac{1}{\sum^k_{i=1} c_i}
		\sum\limits_{i=1}^k \sum\limits_{j=1}^{c_i} 
		\frac{\left( m_{i,j} - \widehat{m}_{i,j}\right)^2}{\mathrm{SE}_i^2}
	},
\end{align*}

 where $i$ is the evaluation case index ranging from $1$ to $k$, $j$ is the index of the production data chunk ranging from  $1$ to $c_i$ (number of production chunks in a case $i$), and $m_{i,j}$ and $\widehat{m}_{i,j}$ are respectively realized and estimated performance for case $i$ and chunk $j$. $\mathrm{SE}_i$ is the standard error of the $i$-th evaluation case. The results are shown in Table \ref{tab2}.

\begin{table}[ht!]
	\centering
	\noindent\adjustbox{max width=\textwidth}
	{
		\begin{tabular}{@{}rcccccc@{}}
			\toprule
			& \multicolumn{2}{c}{Accuracy} & \multicolumn{2}{c}{AUROC} & \multicolumn{2}{c}{F$_1$}\\
			& NMAE & NRMSE & NMAE & NRMSE & NMAE & NRMSE \\
			\midrule
                \textbf{TEST SET} & 1.62 & 2.88	& 1.45 & 2.30 & 2.53 & 8.23 \\
			\textbf{RT-mod}   & 2.31 & 5.41 & 1.85 & 4.29 & 2.06 & 4.74 \\
            \textbf{COT}    & 2.10 & 4.31 & - & - & - & - \\ 
			\textbf{ATC}      & 1.58 & 2.79	& 1.90 & 3.52 & 2.97 & 9.06 \\
			\textbf{DOC}      & 1.13 & 1.80	& 1.37 & 2.75 & 2.56 & 8.52 \\
			\textbf{CBPE}     & 1.08 & 1.75 & 1.07 & 1.68 & 1.03 & 2.12 \\
			\textbf{IW}       & 1.04 & 1.40 & 1.06 & 1.56 & 1.07 & 1.74 \\
			\textbf{PAPE}    & \textbf{0.97} & \textbf{1.28} & \textbf{0.99} & \textbf{1.45} & \textbf{0.90} & \textbf{1.34} \\
			\bottomrule
			\vspace{0.3pt}
		\end{tabular}
	}
	\caption{
		NMAE and NRMSE of the evaluated methods for each estimated metric.
	}\label{tab2}
\end{table}

PAPE, together with CBPE and IW, shows strong improvement over the TEST SET baseline. PAPE significantly outperforms the second-best method for each estimation and evaluation metric (paired Wilcoxon signed-rank test \(p = 0.0\)). PAPE result of NMAE equal to 0.97 for accuracy estimation means that the estimation is, on average, less than 1 SE away from the realized performance. For intuition, the NMAE of PAPE for the evaluation case shown in Figure \ref{fig:expsetup} is equal to 1.11. The TEST SET performance NMAE for accuracy is 1.68, indicating that accuracy changes significantly in the production data chunks. If the performance was stable with only random normal fluctuations, the NMAE of the TEST SET performance would be around 0.8\footnote{That is because the average absolute deviation for a normal distribution is \(\sqrt{\frac{2}{\pi}}\approx 0.8\) of its standard deviation.}. The changes for F$_1$ are much stronger, as shown by the increased NMAE of TEST SET - 2.49. This explains why performance estimation methods provide the strongest improvement compared to the TEST SET performance for this metric. Additionally, we calculated NMAE and NRMSE for each monitored model type - PAPE consistently outperformed IW for each of the monitored models and each estimation and evaluation metric.

\begin{figure}
	\centering
\includegraphics[width=0.32\textwidth]{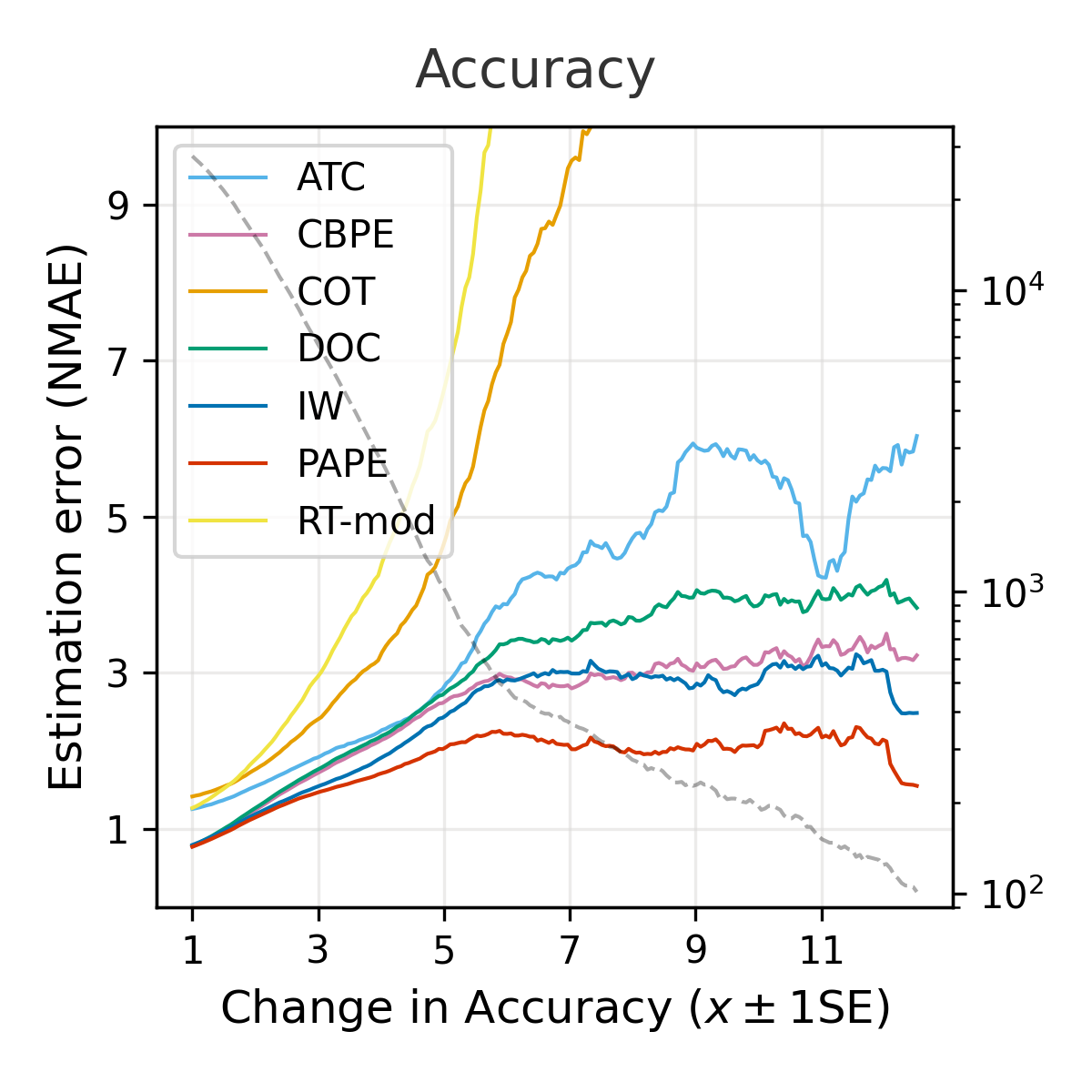}
\includegraphics[width=0.32\textwidth]  {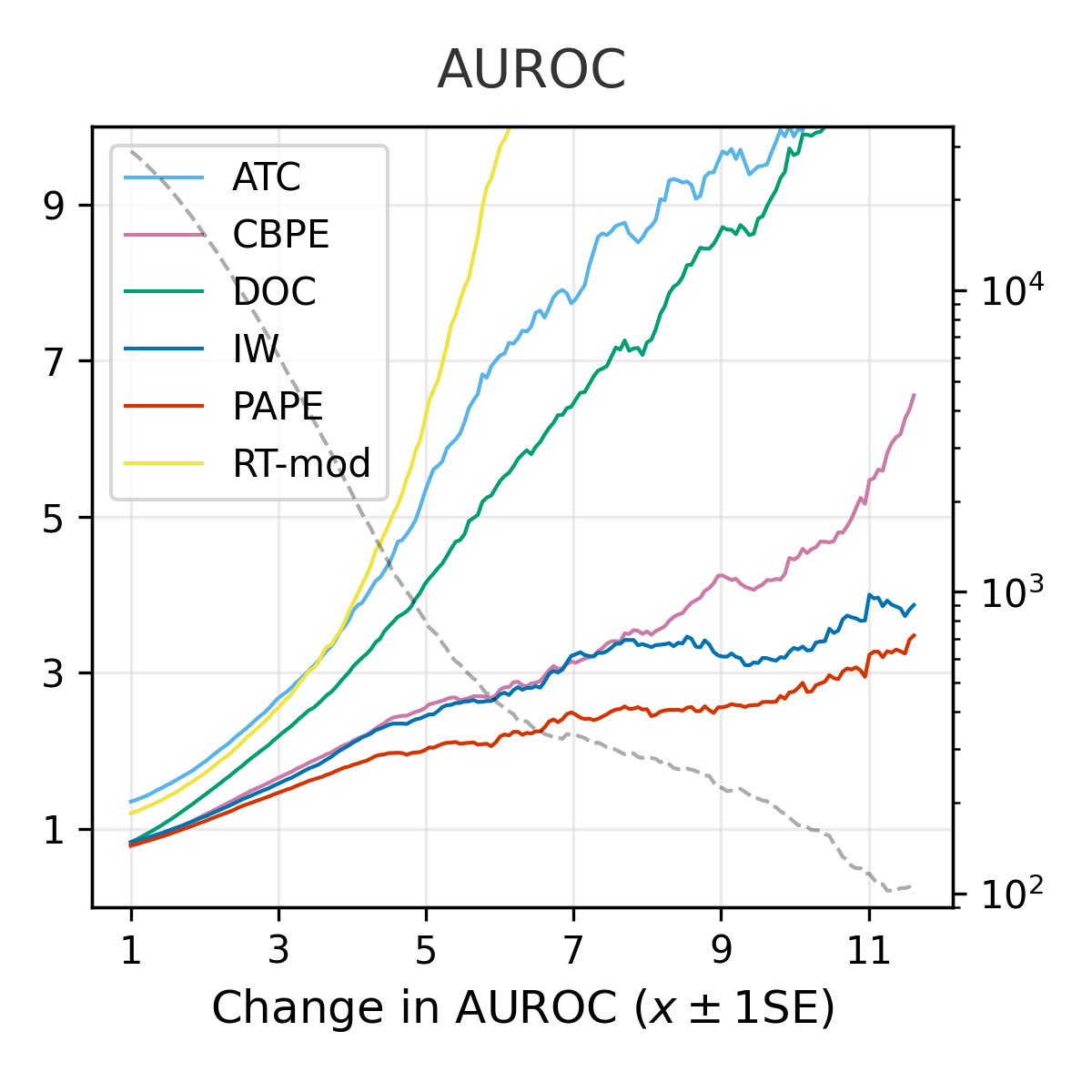}
\includegraphics[width=0.32\textwidth]{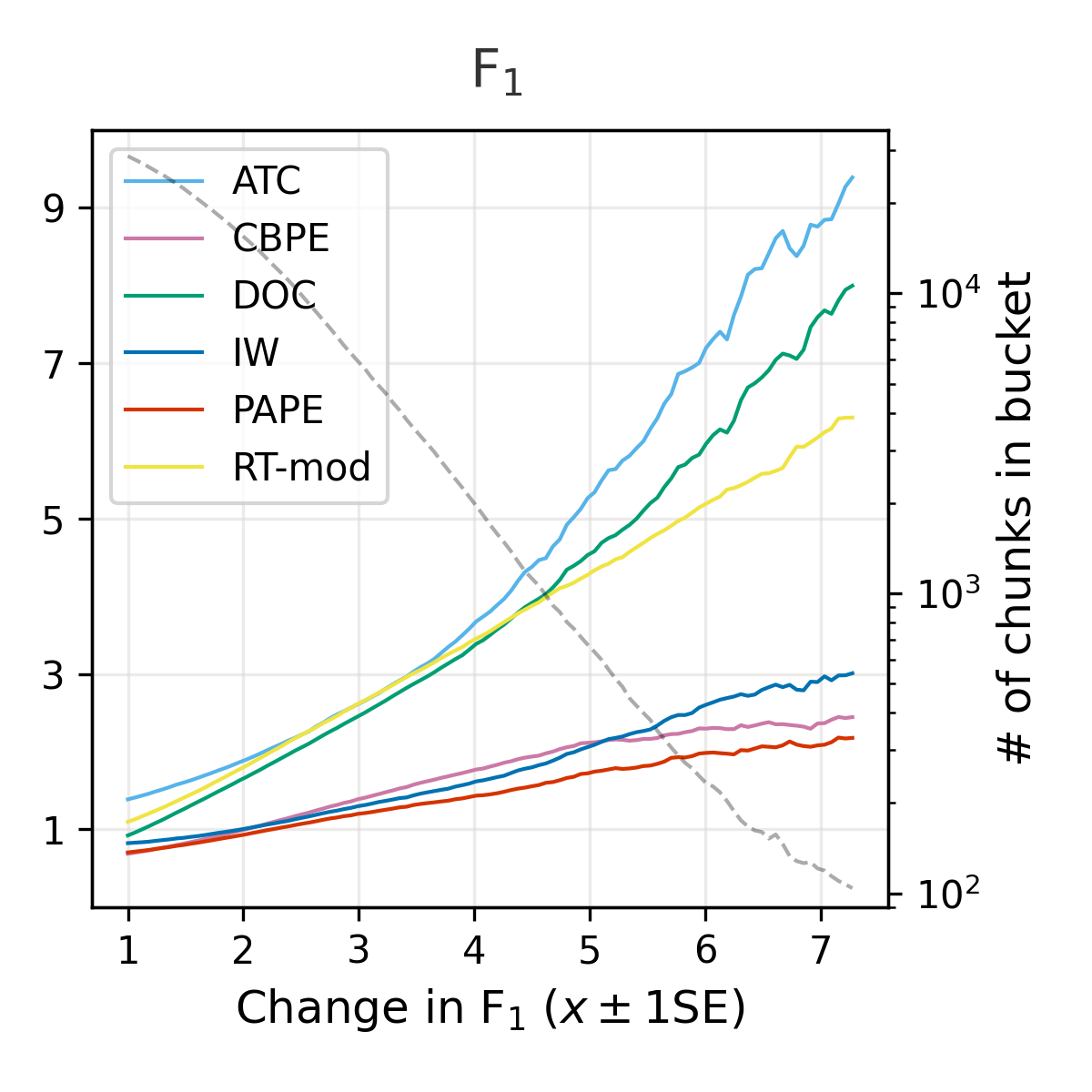}

	\caption{
		Estimation errors (NMAE) of estimated metric vs. realized absolute change as SE for all estimators. The x-axis indicates the center of the data bucket - for example, value 1 indicates a bucket that contains data chunks for which the absolute performance change was between 0 - 2 SE. The left y-axis shows NMAE of the evaluated method for the data bucket. The right y-axis shows the number of data chunks in each bucket on a logarithmic scale as depicted by the grey dashed line.
	}
	\label{fig:r1}
\end{figure}

We also analyzed each performance estimation algorithm's accuracy across different levels of realized performance changes. For relatively small realized performance changes (within $\pm 2$SE), the TEST SET performance baseline is sufficiently accurate. Any useful performance estimator should outperform this baseline at least when the realized performance changes are significant (not within $\pm 2$SE). We sorted the data by the magnitude of the performance change and calculated the rolling NMAE for 2SE-wide intervals. The results are shown in Figure \ref{fig:r1}.

For all the estimated metrics, PAPE shows the best estimation quality in nearly the whole evaluated region. It gets significantly better than other methods when the change in realized performance is large. IW and CBPE show comparable performance, with IW being more accurate in most buckets. The RT and COT methods produce estimation errors, which render them unusable for all metrics used in the experiment. The DOC and ATC methods are somewhat competitive when estimating accuracy (for which they were designed) and when the change in realized accuracy is small enough. For the other metrics, the estimation errors are too high for practical monitoring purposes.

\section{Discussion}\label{sec:discussion}

PAPE is perhaps best understood as an improvement over the CBPE method~\cite{kivimaki:2025b}. PAPE retains all of the benefits of CBPE, such as applicability to multiple metrics instead of just accuracy, but also addresses the problem of deteriorating calibration under covariate shift, which is the major shortcoming of CBPE. Our experimental findings show that PAPE consistently outperforms CBPE in all experiments. Alternatively, PAPE can be viewed as an in-between solution, trying to achieve the best of both worlds of IW and multicalibration approaches. 

Multicalibration requires post-processing, where the original model is iteratively updated through an auditing process~\cite{globus:2023}. This might result in predictions different from those of the original model. Also, the hypothesis space $\mathcal{H}$ needs to be large enough to contain $w_{s\rightarrow t}$, or at least good approximations of them, for any potential shifts. Unfortunately, especially in high-dimensional settings, this might be infeasible or at least increase the computational burden significantly. In contrast, PAPE focuses on one target distribution at a time, learning the density ratios required directly from the data. Thus, PAPE does not need to compromise performance by trying to deal with multiple distributions at the same time. PAPE is also non-invasive, making no alterations to the monitored model, which makes it more suitable for monitoring purposes. In fact, the model being monitored does not need to be calibrated at all, since PAPE provides calibration on the fly. 

On the other hand, IW is known to suffer from high variance estimates when there is a significant discrepancy between
the source and target distributions~\cite{li:2020}. Since the labels used with IW are either 0 or 1, it is also susceptible to large sampling errors, especially in small sample settings that are typical in model monitoring. Since PAPE uses soft scores from the interval $[0, 1]$, it tends to smooth out these sampling effects. We provide a more comprehensive comparative analysis of the variances of PAPE and IW in Appendix~\ref{app:variance_comparison}.

\subsection{Limitations}\label{sec:limitations}

PAPE will not work under \textit{concept shift}, that is, if $p_s(y|\boldsymbol{x})\neq p_t(y|\boldsymbol{x})$. Also, when operating in a covariate shift setting, the data may shift outside the support of the source distribution, which means that there is a region $S \subseteq \mathcal{X}$, where for all $\boldsymbol{x}\in S$ we have $p_t(\boldsymbol{x})>0$ and $p_s(\boldsymbol{x})=0$. Weighted calibration for instances from such regions is not possible as the weights $w_{s \rightarrow t}(\boldsymbol{x})$ are not defined. 

PAPE hinges on calibration performance within the target distribution $\mathcal{D}_t$, which in turn depends on good enough density ratio estimates. Both can be hard to achieve if not enough labeled reference data is available to train the DRE model and the calibration mapping. This data demand increases with the dimensionality of the covariates. As such, PAPE is most performant with low-dimensional data.

\subsection{Future Work}\label{future work}

Although we have described PAPE as a method for estimating the performance of binary classifiers, extending it to multiclass classifiers is straightforward. For instance, in the case of macro-averaged metrics - where performance is computed separately for each class in a one-vs-all manner and then averaged - PAPE can be applied directly by estimating the per-class performance and averaging the results. In fact, PAPE can be used to monitor the performance of any model that produces confidence scores in addition to its predictions. We leave the details of this for future work. 

One challenge with PAPE or any other method relying on density-ratio estimation is that estimating these ratios becomes increasingly hard with high-dimensional data. Examining the sample complexity requirements and relating those to the estimation quality of PAPE is an interesting and important research question to be addressed by later research.

By using AUROC as an estimated metric in our experiments, we expanded the scope of CBPE to a previously unestimated metric. We explain in Appendix~\ref{app:implementation} how this was done precisely. Contrary to CBPE, where a full distribution for each metric is estimated, our approach with AUROC results only in an approximation of the expected value of the metric. We intend to examine metrics such as AUROC and AUPR further to provide a way to estimate the full distributions of these metrics, which require calculations over multiple thresholds.


\section{Conclusion}\label{sec:conclusion}

We introduced PAPE, an innovative approach for accurately estimating the performance of binary classification models when faced with covariate shift. We examined its theoretical properties and provided a bound for its estimation error for composable metrics under approximate calibration. We performed rigorous testing for PAPE using real-world datasets drawn from US Census data, introducing novel evaluation metrics essential for aggregating results over multiple datasets and model monitoring scenarios. We analyzed over 900 model-dataset pairs and generated more than 36,000 data chunks for thorough evaluation. The results demonstrated that PAPE significantly outperforms existing methods across all metrics assessed.


\begin{ack}

This work was partly supported by local authorities (“Business Finland”) under grant agreement 23004 ELFMo of the ITEA4 programme, which funded one of the authors. The remaining research was conducted at NannyML - a venture-backed open-source software company focused on post-deployment data science solutions - which has received public R\&D funding from Flanders Innovation \& Entrepreneurship (VLAIO) under project number HBC.2022.0846.
\end{ack}

\newpage
\begin{CJK*}{UTF8}{gbsn}
\bibliography{references}
\end{CJK*}

\newpage

\appendix

\section{Implementation Details}\label{app:implementation}

In this section, we first explain how F$_1$ and AUROC were estimated for PAPE (and CBPE) in the experiments in Section~\ref{sec:experiments}. Then, we describe our implementation for estimating the same metrics with ATC and DOC, since these methods were not originally intended for estimating any other metrics besides accuracy. 

\subsection{Estimating F$_1$ and AUROC with PAPE}

We omitted confidence intervals in our experiments to relieve the computational burden and because only CBPE and PAPE are capable of providing these intervals. Thus, with CBPE and PAPE, we resorted to deriving only point estimates. Let us recall equation \eqref{eq:MCBPE}:
\begin{equation*}
	\widehat{m}_{(c \circ f, g, \mathcal{D}_t)} = 
	\mathbb{E}_{p_t(\boldsymbol{x})}
	\left[ 
	c(f(\boldsymbol{x}))m\left(\hat{y},1\right) +
	\left(1 - c(f(\boldsymbol{x}))\right)m\left(\hat{y},0\right)
	\right].
\end{equation*}
One can use the sample mean to approximate this expectation and use the approximation in turn to estimate the performance of \textit{composable} metrics, such that can be calculated as a mean of observation-level values, like accuracy. For other metrics, we start by estimating elements of the confusion matrix. Assume we have a sample of $n$ instances and that there are $n_+$ positive and $n_-$ negative predictions in the sample. Then, as explained in~\cite{kivimaki:2025b}, the elements of the confusion matrix can be estimated with:
\begin{align*}
	\widehat{TP} & = \frac{1}{n_+}\sum_{i=1}^n c\bigl(f(\boldsymbol{x_i})\bigr) \cdot \hat{y}_i \\
    \widehat{FP} & = \frac{1}{n_+}\sum_{i=1}^n \Big(1-c\bigl(f(\boldsymbol{x_i})\bigr)\Big) \cdot \hat{y}_i \\
	\widehat{TN} & = \frac{1}{n_-}\sum_{i=1}^n c\bigl(f(\boldsymbol{x_i})\bigr) \cdot (1-\hat{y}_i) \\
	\widehat{TN} & = \frac{1}{n_-}\sum_{i=1}^n \Big(1-c\bigl(f(\boldsymbol{x_i})\bigr)\Big) \cdot (1-\hat{y}_i) \\
\end{align*}

With the estimated elements of the confusion matrix, one can estimate F$_1$ with:
\begin{equation}\label{eq:f1}
	\widehat{\mathrm{F}_1} \approx \frac{2\cdot\widehat{TP}}{\widehat{TP} + \widehat{FN} + n_+}
\end{equation}
If we treat the value of F$_1$ as a random variable $X_{F_1}$, the approximation given in~(\ref{eq:f1}) converges to $\mathbb{E}[X_{F_1}]$ rapidly when $n$ grows. This expectation, in turn, is an unbiased and consistent estimator of F$_1$ under perfect calibration.~\cite{kivimaki:2025b}

To estimate AUROC, the model's binary predictions $\hat{y}$ are not needed. We use the scores $f(x)$ returned by the classifier to define a new thresholding function $h_v$:
\begin{equation*}
	h_v\bigl(f(\boldsymbol{x}), v\bigr) = 
	\begin{cases}
		1,	& f(x) \geq v \\
		0,	& f(x) < v
	\end{cases}
\end{equation*}
For each $v\in R(f)$ and for each $\boldsymbol{x}_i$ in the sample, we can use
$\hat{y}_i=h_v(f(\boldsymbol{x}_i), v)$ and estimate the elements of the confusion matrix as explained above. Then, using these estimates, we can approximate true positive rates (TPR) and false positive rates (FPR) as 
\begin{align*}
    \widehat{TPR}_v &\approx \frac{\widehat{TP}_v}{\widehat{TP}_v + \widehat{FN}_v} \\
    \widehat{FPR}_v &\approx \frac{\widehat{TN}_v}{\widehat{TN}_v + \widehat{FP}_v}    
\end{align*}
Finally, we can calculate an estimate for AUROC with the approximated TPR and FPR. In our experiments, we tried several different calibration mappings and settled on LGBM, since it gave the best overall performance.

\subsection{Estimating F$_1$ and AUROC with Other Estimators}
Since COT~\cite{lu:2023} is essentially estimating the 0/1 classification error, there is no clear way of how it could be used to estimate metrics other than accuracy. For this reason, we left it out of experiments where F$_1$ and AUROC were estimated. Although ATC~\cite{garg:2022} and DOC~\cite{guillory:2021} were also originally designed for estimating only accuracy, we used the following procedures to expand them for estimating other metrics as well.

When estimating the F$_1$ score and AUROC with ATC, we applied the same logic as with accuracy. For accuracy, ATC finds a confidence threshold such that the fraction of instances from the reference data set with predicted confidence score above the threshold matches model accuracy in the reference set. With production data, the fraction of confidence scores above the same confidence threshold is taken as the estimated accuracy for the production data. When estimating the F$_1$ and AUROC, we similarly found confidence thresholds matching these metrics with reference data and used the fraction of production data instances with confidence scores above those thresholds as estimates.

With DOC, we simulated distribution shifts by resampling reference data, as we did with accuracy. We collected the F$_1$ and AUROCs and fitted a linear regression model with the difference of average confidence between the reference and simulated sets as the single feature and the metric of interest as the target (just as with accuracy). We did not apply calibration to the confidence scores for either the DOC or ATC methods. Our reasoning behind this is that with DOC, the linear regression model is indifferent with respect to any rescaling of the confidence scores. On the other hand, with ATC, we found that the reported differences between the calibrated and uncalibrated versions of ATC with binary classification in~\cite{kivimaki:2025} using the models we included in our experiments were so minimal that we chose to include only the variant that had a better reported overall performance, in this case, the uncalibrated version.

\section{Proofs}\label{app:proofs}
In this section, we provide proofs for the theorems presented in the main paper. We begin with Theorem~\ref{thr:calibration}.

\subsection{Proof of Theorem~\ref{thr:calibration}}\label{app:proofs:th3_1}
To improve readability, we start by proving two lemmas, which we can then leverage in the main proof\footnote{Acknowledgement: Our derivations follow closely those of Prof. Aaron Roth on his \href{https://uncertaintyclass.com/}{"Learning with Conditional Guarantees"} course, which he teaches at the University of Pennsylvania.}. First, we make use of the following connection between the expectations of the source $\mathcal{D}_s = p_s(\boldsymbol{x}, y)$ and target $\mathcal{D}_t = p_t(\boldsymbol{x}, y)$ distributions:
\begin{lemma}\label{thr:lemma1}
    Assume $p_s(y|\boldsymbol{x}) = p_t(y|\boldsymbol{x})$ and fix any $S \subseteq \mathcal{X}$. Then, for any integrable function $F: \mathcal{X} \times \mathcal{Y} \rightarrow \mathbb{R}$, we have:
\begin{equation*}
        P_{\mathcal{D}_s}(\boldsymbol{x} \in S)\mathbb{E}_{\mathcal{D}_s}[w_{s \rightarrow t}(\boldsymbol{x}) \cdot F(\boldsymbol{x}, y)\mid \boldsymbol{x} \in S] = P_{\mathcal{D}_t}(\boldsymbol{x} \in S)\mathbb{E}_{\mathcal{D}_t}[F(\boldsymbol{x}, y)\mid \boldsymbol{x} \in S].
    \end{equation*}
\end{lemma}
\begin{proof}
\begin{align*}
    &P_{\mathcal{D}_s}(\boldsymbol{x} \in S)\mathbb{E}_{\mathcal{D}_s}[w_{s \rightarrow t} \cdot F(\boldsymbol{x}, y)\mid \boldsymbol{x} \in S] \\
    = &\int_{\boldsymbol{x} \in S} p_s(\boldsymbol{x})\cdot w_{s \rightarrow t}(\boldsymbol{x})\cdot \mathbb{E}_{p(y|\boldsymbol{x})}[F(\boldsymbol{x},y)]~d\boldsymbol{x}\\
    = &\int_{\boldsymbol{x} \in S} p_s(\boldsymbol{x})\cdot \frac{p_t(\boldsymbol{x})}{p_s(\boldsymbol{x})} \cdot \mathbb{E}_{p(y|\boldsymbol{x})}[F(\boldsymbol{x},y)]~d\boldsymbol{x}\\
    = &\int_{\boldsymbol{x} \in S} p_t(\boldsymbol{x}) \cdot \mathbb{E}_{p(y|\boldsymbol{x})}[F(\boldsymbol{x},y)]~d\boldsymbol{x}\\
    = &P_{\mathcal{D}_t}(\boldsymbol{x} \in S)\mathbb{E}_{\mathcal{D}_t}[F(\boldsymbol{x}, y)\mid \boldsymbol{x} \in S].
\end{align*}
\end{proof}

Now, we can leverage Lemma~\ref{thr:lemma1} to prove the following:
\begin{lemma}\label{thr:lemma2}
    Assume $p_s(y|\boldsymbol{x}) = p_t(y|\boldsymbol{x})$. Suppose f is $\alpha$-approximately multicalibrated in $\mathcal{D}_s$ with respect to $\mathcal{H}$. Then, $f$ is also $\alpha$-approximately multicalibrated in $\mathcal{D}_t$ with respect to $\mathcal{H}_{s \rightarrow t}$, where 
    \begin{equation*}
        \mathcal{H}_{s \rightarrow t} = \left\{\frac{h(\boldsymbol{x})}{w_{s \rightarrow t}(\boldsymbol{x})}:h(\boldsymbol{x})\in \mathcal{H}\right\}.
    \end{equation*}
\end{lemma}
\begin{proof}
    Since $f$ is $alpha$-approximately multicalibrated in $\mathcal{D}_s$ with respect to $\mathcal{H}$, for every $h \in \mathcal{H}$, we have:
    \begin{align*}
        \alpha &\ge K(f, \mathcal{D}_s, h) \\
        &= \sum_{v \in R(f)}P_{\mathcal{D}_s}(f(\boldsymbol{x})=v) \bigg|\mathbb{E}_{\mathcal{D}}[h(\boldsymbol{x})(y - v) \mid f(\boldsymbol{x})=v]\bigg| \\
        &= \sum_{v \in R(f)} \bigg|P_{\mathcal{D}_s}(f(\boldsymbol{x})=v) \cdot\mathbb{E}_{\mathcal{D}_s}[h(\boldsymbol{x})(y - v) \mid f(\boldsymbol{x})=v]\bigg| \\
        &= \sum_{v \in R(f)} \bigg|P_{\mathcal{D}_s}(f(\boldsymbol{x})=v) \cdot\mathbb{E}_{\mathcal{D}_s}\left[w_{s \rightarrow t}(\boldsymbol{x})\cdot\frac{h(\boldsymbol{x})}{w_{s \rightarrow t}(\boldsymbol{x})}\cdot(y - v)~\Big|~f(\boldsymbol{x})=v\right]\bigg| \\
        &= \sum_{v \in R(f)} \bigg|P_{\mathcal{D}_t}(f(\boldsymbol{x})=v) \cdot\mathbb{E}_{\mathcal{D}_t}\left[\frac{h(\boldsymbol{x})}{w_{s \rightarrow t}(\boldsymbol{x})}(y - v)~\Big|~f(\boldsymbol{x})=v\right]\bigg| \\
        &= K\left(f, \mathcal{D}_t, \frac{h}{w_{s \rightarrow t}}\right),
    \end{align*}
    where Lemma~\ref{thr:lemma1} is applied to each of the terms
    \begin{equation*}
        P_{\mathcal{D}_s}(f(\boldsymbol{x})=v) \cdot\mathbb{E}_{\mathcal{D}_s}\left[w_{s \rightarrow t}(\boldsymbol{x})\cdot\frac{h(\boldsymbol{x})}{w_{s \rightarrow t}(\boldsymbol{x})}\cdot(y - v)~\Big|~f(\boldsymbol{x})=v\right]
    \end{equation*}
    using $S=\{\boldsymbol{x}:f(\boldsymbol{x})=v\}$ and $F(\boldsymbol{x}, y)=\frac{h(\boldsymbol{x})}{w_{s \rightarrow t}(\boldsymbol{x})}(y - v)$.
\end{proof}
We are now ready to prove Theorem~\ref{thr:calibration}, which we will restate here:

\begin{reptheorem}{thr:calibration}
  Assume that $p_s(y|\boldsymbol{x}) = p_t(y|\boldsymbol{x})$ and that $f$ is $\alpha$-approximately multicalibrated in $\mathcal{D}_s$ with respect to $\mathcal{H}$. If $w_{s \rightarrow t} \in \mathcal{H}$, then $K(f, \mathcal{D}_t) \le \alpha$.
\end{reptheorem}
\begin{proof}
    Since we assumed that $w_{s \rightarrow t} \in \mathcal{H}$, we can choose $h=w_{s \rightarrow t}$ and apply Lemma~\ref{thr:lemma2} to get
    \begin{align*}
        \alpha &\ge K\left(f, \mathcal{D}_t, \frac{h}{w_{s \rightarrow t}}\right) \\
        &= \sum_{v \in R(f)} \bigg|P_{\mathcal{D}_t}(f(\boldsymbol{x})=v) \cdot\mathbb{E}_{\mathcal{D}_t}\left[\frac{h(\boldsymbol{x})}{w_{s \rightarrow t}(\boldsymbol{x})}(y - v)~\Big|~f(\boldsymbol{x})=v\right]\bigg| \\
        &=\sum_{v \in R(f)} P_{\mathcal{D}_t}(f(\boldsymbol{x})=v) \bigg|\mathbb{E}_{\mathcal{D}_t}\left[(y - v)~\big|~f(\boldsymbol{x})=v\right]\bigg| \\
        &= K(f, \mathcal{D}_t).
    \end{align*}
\end{proof}

\subsection{Proof of Theorem~\ref{thr:estimation}}\label{app:proofs:th3_2}

In this section, we will offer the proof of Theorem~\ref{thr:estimation}, which we restate below:

\begin{reptheorem}{thr:estimation}
    Let $c \circ f$ be $\alpha$-calibrated in $\mathcal{D}_t$. Also, assume that $f$ has a countable image set $R(f)$, $m$ is a composable metric with $0 \le m(\hat{y}, y) \le 1$, and that $\hat{m}$ is its PAPE estimate. Then,
    \begin{equation*}
    |m_{\left(f, g, \mathcal{D}_t\right)}\ - \widehat{m}_{(c \circ f, g, \mathcal{D}_t)}| \le \alpha.
    \end{equation*}
\end{reptheorem}
\begin{proof}
We start by decomposing the calibration error as a sum of terms related to each levelset $\{\boldsymbol{x} \in \mathcal{X}: f(\boldsymbol{x})=v\}$. For each $v$, we assume that a calibration error $\alpha_v$ remains after applying the calibration mapping $c$. Formally,
\begin{equation*}
    \mathbb{E}_{\mathcal{D}_t}[y \mid f(\boldsymbol{x})=v]=P_{\mathcal{D}_t}\left[y=1 \mid f(\boldsymbol{x})=v\right] = c\left(v\right)+\alpha_v 
	\label{eq:approx_calib}.
\end{equation*}
For any fixed $v$ and deterministic $c$, clearly $\mathbb{E}_{\mathcal{D}_t}[c(v) \mid f(\boldsymbol{x})=v] = c(v)$, which yields us
\begin{align*}
    \mathbb{E}_{\mathcal{D}_t}[y \mid f(\boldsymbol{x})=v] &= c\left(v\right)+\alpha_v \\
    \mathbb{E}_{\mathcal{D}_t}[y \mid f(\boldsymbol{x})=v] - c\left(v\right) &= \alpha_v \\
    \mathbb{E}_{\mathcal{D}_t}[y - c\left(v\right) \mid f(\boldsymbol{x})=v] &= \alpha_v.
\end{align*}
The total calibration error is then:
\begin{equation*}
       \sum_{v\in R(f)}P_{p_t(\boldsymbol{x})}(f(\boldsymbol{x})=v)\bigg|\mathbb{E}_{\mathcal{D}_t}[y-c(v)|f(\boldsymbol{x})=v]\bigg| = \sum_{v\in R(f)}P_{p_t(\boldsymbol{x})}(f(\boldsymbol{x})=v)\left|\alpha_v\right| \le \alpha.
       \label{eq:expected_calib}
\end{equation*}
Furthermore, notice that since the levelsets $\{\boldsymbol{x} \in \mathcal{X}: f(\boldsymbol{x})=v\}$ form a partition of $\mathcal{X}$, for any integrable function $F: [0, 1] \rightarrow \mathbb{R}$ we have $\mathbb{E}_{p_t(\boldsymbol{x})}[F(v)]=\sum_{v \in R(f)} P_{p_t(\boldsymbol{x})}(f(\boldsymbol{x})=v)F(v)$. Thus, we can write:
\begin{align*}
    m_{(f, g, \mathcal{D}_t)} = &~\mathbb{E}_{\mathcal{D}_t} \left[m\left(g(f(\boldsymbol{x})), y\right)\right]\\
      = &\sum_{v \in R(f)} P_{p_t(\boldsymbol{x})}\big(f(\boldsymbol{x})=v\big) \mathbb{E}_{\mathcal{D}_t}\big[m\big(g(f(\boldsymbol{x})), y\big) \mid f(\boldsymbol{x})=v\big]\\
      = &\sum_{v \in R(f)} P_{p_t(\boldsymbol{x})}\big(f(\boldsymbol{x})=v\big) \Big(m\big(g(v), 1\big)\cdot P_{\mathcal{D}_t}\big(y=1 \mid f(\boldsymbol{x})=v\big)~+ \\
      &\hspace{11em} m\big(g(v), 0\big)\cdot P_{\mathcal{D}_t}\big(y=0 \mid f(\boldsymbol{x})=v\big)\Big) \\
      = &\sum_{v \in R(f)} P_{p_t(\boldsymbol{x})}\big(f(\boldsymbol{x})=v\big) \Big(m\big(g(v), 1\big)\big(c(v)+\alpha_v\big) + m\big(g(v), 0\big)\big(1-(c(v)+\alpha_v)\big)\Big)\\
      = &\sum_{v \in R(f)} P_{p_t(\boldsymbol{x})}\big(f(\boldsymbol{x})=v\big) \Big(m\big(g(v), 1\big)c(v) + m\big(g(v), 0\big)\big(1-c(v)\big)\Big)~+\\
      & \sum_{v \in R(f)} P_{p_t(\boldsymbol{x})}\big(f(\boldsymbol{x})=v\big) \Big(\alpha_v\big(m(g(v), 1)  - m(g(v), 0)\big)\Big)\\
      = &~\widehat{m}_{(f, g, \mathcal{D}_t)} + \sum_{v \in R(f)} P_{p_t(\boldsymbol{x})}\big(f(\boldsymbol{x})=v\big) \Big(\alpha_v\big(m(g(v), 1)  - m(g(v), 0)\big)\Big),  
\end{align*}
which we can further manipulate as
\begin{equation}\label{eq:estimation_error}
    m_{(f, g, \mathcal{D}_t)} - \widehat{m}_{(f, g, \mathcal{D}_t)} =  \sum_{v \in R(f)} P_{p_t(\boldsymbol{x})}\big(f(\boldsymbol{x})=v\big) \Big(\alpha_v\big(m(g(v), 1)  - m(g(v), 0)\big)\Big). 
\end{equation}
For any metric $m$ with $0 \le m(\hat{y}, y) \le 1$, we have
\begin{align*}
    |m(g(v), 1) - m(g(v), 0)| &\le 1 \\
    |\alpha_v(m(g(v), 1)  - m(g(v), 0))| &\le |\alpha_v|,
\end{align*}
Thus, if we take the absolute values of both sides of Equation~\ref{eq:estimation_error}, we get
\begin{align*}
    |m_{(f, g, \mathcal{D}_t)} - \widehat{m}_{(f, g, \mathcal{D}_t)}| &= \left| \sum_{v \in R(f)} P_{p_t(\boldsymbol{x})}\big(f(\boldsymbol{x})=v\big) \Big(\alpha_v\big(m(g(v), 1)  - m(g(v), 0)\big)\Big)\right| \\
    &= \sum_{v \in R(f)} P_{p_t(\boldsymbol{x})}\big(f(\boldsymbol{x})=v\big) \cdot \left|\alpha_v\big(m(g(v), 1)  - m(g(v), 0)\big)\right| \\
    &\le \sum_{v \in R(f)} P_{p_t(\boldsymbol{x})}\big(f(\boldsymbol{x})=v\big) |\alpha_v| \\
    &\le \alpha,
\end{align*}
completing the proof.
\end{proof}

\subsection{Estimation Error Under Imperfect Weights}\label{app:proofs:imperfect_weights}
Our proof of Theorem~\ref{thr:calibration} assumes that we have access to exact density ratios. Here, we will produce an upper bound for the calibration error when we resort to approximating the weights with some function $h \in \mathcal{H}$. To quantify the approximation error, we make use of the following definition:
\begin{definition}
    Assume that $p_s(y|\boldsymbol{x}) = p_t(y|\boldsymbol{x})$. For a function $h: \mathcal{X} \rightarrow \mathbb{R}$, we write
    \begin{equation*}
        \epsilon(h, w_{s \rightarrow t}) = \mathbb{E}_{p_s(\boldsymbol{x})}[|h(\boldsymbol{x})-w_{s\rightarrow t}(\boldsymbol{x})|].
    \end{equation*}
    Similarly, for any subset $S\subseteq\mathcal{X}$, we write:
    \begin{equation*}
        \epsilon(h, w_{s \rightarrow t}, S) = \mathbb{E}_{p_s(\boldsymbol{x})}[|h(\boldsymbol{x})-w_{s\rightarrow t}(\boldsymbol{x})| \mid \boldsymbol{x} \in S].
    \end{equation*}
\end{definition}

Now, we can prove the following lemma:
\begin{lemma}\label{thr:lemma3}
    Assume $p_s(y|\boldsymbol{x}) = p_t(y|\boldsymbol{x})$ and fix any $S \subseteq \mathcal{X}$. Then, for any integrable functions $F: \mathcal{X} \times \mathcal{Y} \rightarrow [-1, 1]$ and $h:\mathcal{X} \rightarrow \mathbb{R}$, we have:
    \begin{align*}        
        &\big|P_{\mathcal{D}_s}(\boldsymbol{x} \in S)\cdot \mathbb{E}_{\mathcal{D}_s}[h(\boldsymbol{x}) \cdot F(\boldsymbol{x}, y)\mid \boldsymbol{x} \in S]\big| \\
        \ge~&\big|P_{\mathcal{D}_t}(\boldsymbol{x} \in S)\cdot \mathbb{E}_{\mathcal{D}_t}[F(\boldsymbol{x}, y)\mid \boldsymbol{x} \in S] \big| - P_{\mathcal{D}_s}(\boldsymbol{x} \in S)\cdot  \epsilon(h, w_{s \rightarrow t}, S).        
    \end{align*}
\end{lemma}
\begin{proof}
    We can use Lemma~\ref{thr:lemma1} to write
    \begin{align*}
        &\big| P_{\mathcal{D}_s}(\boldsymbol{x} \in S)\cdot \mathbb{E}_{\mathcal{D}_s}[h(\boldsymbol{x}) \cdot F(\boldsymbol{x}, y)\mid \boldsymbol{x} \in S] -
        P_{\mathcal{D}_t}(\boldsymbol{x} \in S)\cdot \mathbb{E}_{\mathcal{D}_t}[F(\boldsymbol{x}, y)\mid \boldsymbol{x} \in S] \big| \\        
        =~&\big| P_{\mathcal{D}_s}(\boldsymbol{x} \in S)\mathbb{E}_{\mathcal{D}_s}[h(\boldsymbol{x}) \cdot F(\boldsymbol{x}, y)\mid \boldsymbol{x} \in S] -
        P_{\mathcal{D}_s}(\boldsymbol{x} \in S)\mathbb{E}_{\mathcal{D}_s}[w_{s \rightarrow t}(\boldsymbol{x})F(\boldsymbol{x}, y)\mid \boldsymbol{x} \in S] \big| \\        
        =~&P_{\mathcal{D}_s}(\boldsymbol{x} \in S)\cdot\big|\mathbb{E}_{\mathcal{D}_s}[(h(\boldsymbol{x})-w_{s \rightarrow t}(\boldsymbol{x})) \cdot F(\boldsymbol{x}, y)\mid \boldsymbol{x} \in S]  \big| \\        
        \le~&P_{\mathcal{D}_s}(\boldsymbol{x} \in S)\cdot \max_{(\boldsymbol{x},y)\in (\mathcal{X},\mathcal{Y})}|F(\boldsymbol{x},y)|\cdot \mathbb{E}_{p_s(\boldsymbol{x}}[|h(\boldsymbol{x})-w_{s \rightarrow t}(\boldsymbol{x})| \mid \boldsymbol{x} \in S] \\    
        \le~&P_{\mathcal{D}_s}(\boldsymbol{x} \in S)\cdot  \epsilon(h, w_{s \rightarrow t}, S),
    \end{align*}
    since clearly $\underset{(\boldsymbol{x},y)\in (\mathcal{X},\mathcal{Y})}{\max}|F(\boldsymbol{x},y)|\le 1$. Then, we can reverse the direction and use the reverse triangle inequality to write 
    \begin{align*}
        &P_{\mathcal{D}_s}(\boldsymbol{x} \in S)\cdot  \epsilon(h, w_{s \rightarrow t}, S) \\
        \ge~&\big| P_{\mathcal{D}_s}(\boldsymbol{x} \in S)\cdot \mathbb{E}_{\mathcal{D}_s}[h(\boldsymbol{x}) \cdot F(\boldsymbol{x}, y)\mid \boldsymbol{x} \in S] -
        P_{\mathcal{D}_t}(\boldsymbol{x} \in S)\cdot \mathbb{E}_{\mathcal{D}_t}[F(\boldsymbol{x}, y)\mid \boldsymbol{x} \in S] \big| \\
        \ge~&\big|P_{\mathcal{D}_t}(\boldsymbol{x} \in S)\cdot \mathbb{E}_{\mathcal{D}_t}[F(\boldsymbol{x}, y)\mid \boldsymbol{x} \in S] \big| - \big| P_{\mathcal{D}_s}(\boldsymbol{x} \in S)\cdot \mathbb{E}_{\mathcal{D}_s}[h(\boldsymbol{x}) \cdot F(\boldsymbol{x}, y)\mid \boldsymbol{x} \in S]\big|,
    \end{align*}
    from which the statement follows by simply rearranging the terms.
\end{proof}

This lemma can be used to prove the following theorem, which is a relaxed version of Theorem~\ref{thr:calibration} and gives an upper bound for the calibration error with approximate likelihood ratios:
\begin{theorem}\label{thr:cal_relaxed}
  Assume that $p_s(y|\boldsymbol{x}) = p_t(y|\boldsymbol{x})$ and that $f$ is $\alpha$-approximately multicalibrated in $\mathcal{D}_s$ with respect to $\mathcal{H}$. Then, 
  \begin{equation*}
      K(f, \mathcal{D}_t) \le \alpha + \min_{h \in \mathcal{H}}\epsilon(h, w_{s\rightarrow t}).
  \end{equation*}
\end{theorem}
\begin{proof}
    Let $h^*=\underset{h\in\mathcal{H}}{\arg\min}~\epsilon(h, w_{s\rightarrow t})$ and $S_v = \{\boldsymbol{x}\in \mathcal{X}: f(\boldsymbol{x})=v\}$ so that the collection $\{S_v\}_{v\in R(f)}$ forms a partition of $\mathcal{X}$. Thus, by the law of total probability, 
    \begin{equation*}
        \sum_{v \in R(f)} P_{\mathcal{D}_s}(f(\boldsymbol{x})=v)\cdot \epsilon(h^*, w_{s\rightarrow t}, S_v)=\epsilon(h^*, w_{s\rightarrow t}).
    \end{equation*}
    Since $f$ is $\alpha$-approximately calibrated in $\mathcal{D}_s$ with respect to $\mathcal{H}$, we have:
    \begin{align*}
        \alpha \ge~&K(f, \mathcal{D}_t, h^*) \\
        =~&\sum_{v \in R(f)} \bigg|P_{\mathcal{D}_s}(f(\boldsymbol{x})=v) \cdot\mathbb{E}_{\mathcal{D}_s}\left[h^*(\boldsymbol{x})(y - v)~\Big|~f(\boldsymbol{x})=v\right]\bigg| \\
        \ge~&\sum_{v \in R(f)} \bigg|P_{\mathcal{D}_t}(f(\boldsymbol{x})=v) \cdot\mathbb{E}_{\mathcal{D}_t}\left[y - v~\Big|~f(\boldsymbol{x})=v\right]\bigg|~- \\
        &\sum_{v \in R(f)} P_{\mathcal{D}_s}(f(\boldsymbol{x})=v) \cdot \epsilon(h^*, w_{s\rightarrow t}, S_v\}) \\ 
        =~&K(f, \mathcal{D}_t) - \epsilon(h^*, w_{s\rightarrow t}).
    \end{align*}
    Here, Lemma~\ref{thr:lemma3} is applied  to each of the terms  
    \begin{equation*}
        \left|P_{\mathcal{D}_s}(f(\boldsymbol{x})=v) \cdot\mathbb{E}_{\mathcal{D}_s}\left[h^*(\boldsymbol{x})(y - v)~\Big|~f(\boldsymbol{x})=v\right]\right|,
    \end{equation*}
    with $F(\boldsymbol{x}, y) = y-v$.  
\end{proof}

\section{IW and PAPE Variance Comparison}\label{app:variance_comparison}

In this section, we compare the variances of IW and PAPE when used to estimate the accuracy of a model that is trained with data from some source distribution $p_s(\boldsymbol{x}, y)$ in some target distribution $p_t(\boldsymbol{x}, y)$. We operate under the covariate shift assumption of $p_s(y|\boldsymbol{x})=p_t(y|\boldsymbol{x})$. We assume that the scores $S=f(X)$ produced by the classifier are perfectly calibrated in $p_s(\boldsymbol{x}, y)$ (but not necessarily in $p_t(\boldsymbol{x}, y)$). That is, if $(X, Y) \sim p_s(\boldsymbol{x}, y)$, then $P(Y=1\mid S=s)=s\quad \forall s\in [0,1]$. In addition, we assume that there is some discriminator function $g: [0, 1] \rightarrow \{0, 1\}$ mapping the scores to binary predictions $\hat{Y}=g(S)$.

\subsection{Estimators}
We start by describing the estimators in this setting and the notation used. 

\subsubsection{Probabilistic Adaptive Performance Estimation}
The CBPE~\cite{kivimaki:2025b} estimator for accuracy from a sample $(X_1, X_2, ..., X_n)\sim p_t(\boldsymbol{x})^n$ is defined as  
\begin{equation}
    X_{accuracy} = \frac{X_{correct}}{n},
\end{equation}
where $X_{correct}$ follows a Poisson binomial distribution with parameters $Z_i$ defined as
\begin{equation}
    Z_i=
    \begin{cases}
        S_i, & \hat{Y}_i = 1 \\
        1-S_i, & \hat{Y}_i = 0.
    \end{cases}
\end{equation}
Let $s:\mathcal{X} \rightarrow [0, 1]$ denote the mapping $s(X_i)=Z_i$. Under perfect calibration, using $\mathbb{E}_{p_t}[X_{accuracy}]$ as a point estimate yields an unbiased and consistent estimator for accuracy~\cite{kivimaki:2025b}.
\begin{equation}
    \widehat{Acc}_{\text{CBPE}}=\frac{1}{n}\sum_{i=1}^n Z_i.
\end{equation}
However, the assumption of perfect calibration in the source distribution $p_s(\boldsymbol{x}, y)$ does not extend to the target distribution $p_t(\boldsymbol{x}, y)$, which typically results in a biased estimator in any target distribution. PAPE allows us to train a weighted calibrator $c:[0, 1]\rightarrow [0, 1]$ using the same (exact) density ratios $w(\boldsymbol{x})$ as IW (explained in the next section), ensuring that $f^w = c \circ f$ is perfectly calibrated in $p_t(\boldsymbol{x}, y)$. Now we can define 
\begin{equation}
    \widehat{Acc}_{\text{PAPE}}=\frac{1}{n}\sum_{i=1}^n Z_i^w,
\end{equation}
where the difference to $\widehat{Acc}_{\text{CBPE}}$ is that the scores $S_i^w$ used to define $Z_i^w$ originate from $f^w$ instead of $f$. Similarly, we let $s^w:\mathcal{X} \rightarrow [0, 1]$ denote the mapping $s^w(X_i)=Z_i^w$. Under the assumption of exact density ratios $w(\boldsymbol{x})$, $\widehat{Acc}_{\text{PAPE}}$ is unbiased and consistent in the target distribution. 

\subsubsection{Importance Weighting}
The empirical importance-weighted (IW) estimator from a sample $(X_1, X_2, ..., X_n) \sim p_t(\boldsymbol{x})^n$ is defined as  
\begin{equation}
    \widehat{Acc}_{\text{IW}}=\frac{1}{n}\sum_{i=1}^n I_i\,w(\boldsymbol{x}_i),
\end{equation}
where the indicator $I_i$ is defined as
\begin{equation}
    I_i=
    \begin{cases}
        1, & \hat{Y}_i = Y_i \\
        0, & \hat{Y}_i \neq Y_i,
    \end{cases}
\end{equation}
with $\hat{Y_i}=g(f(X_i))$, $\{Y_i\}_{i=1}^n \sim p_s(y \mid \boldsymbol{x})^n$, and for each $X_i = \boldsymbol{x}_i$
\begin{equation}
    w(\boldsymbol{x}_i)=\frac{p_t(\boldsymbol{x}_i)}{p_s(\boldsymbol{x}_i)},   
\end{equation}
is the density ratio relating the marginal source and target distributions, $p_s(\boldsymbol{x})$ and $p_t(\boldsymbol{x})$ respectively (with $p_s(\boldsymbol{x})>0$ whenever $p_t(\boldsymbol{x})>0$). Under the assumption of perfect calibration (in the source distribution), we have $I_i \sim \operatorname{Bernoulli}(Z_i)$. The empirical importance-weighted estimator (under exact density ratios) is known to be unbiased and consistent in the target distribution. 

\subsection{Variance Comparison}
If we have access to exact density ratios, both PAPE and IW are unbiased and consistent estimators for accuracy. Then, the choice between the two should depend only on their sample efficiencies. That is, which estimator has the smallest variance? On the other hand, if density ratios are not exact, it is a reasonable assumption (by the principle of insufficient reason) that both estimators would suffer roughly equally on average.

By using $p_t$ as a shorthand for $X \sim p_t(\boldsymbol{x})$, the per-observation variance of PAPE is, by definition\footnote{In expectations such as $\mathbb{E}_{p_t}[s^w(\boldsymbol{x})]$, we use $\boldsymbol{x}$ as a dummy variable representing a specific realization of $X$. Thus, any reference to $\boldsymbol{x}$ in these expressions is to be understood as a placeholder that is integrated out. }
\begin{equation}
    \operatorname{Var}_{p_t}(Z^w) = \mathbb{E}_{p_t}[(Z^w)^2] - \mathbb{E}_{p_t}[Z^w]^2 = \mathbb{E}_{p_t}[s^w(\boldsymbol{x})^2] - \mathbb{E}_{p_t}[s^w(\boldsymbol{x})]^2.
\end{equation}

Let us take a look at the per-observation terms of the IW estimator in a similar fashion. We start by denoting $W=Iw(X)$, where recall that $I$ is a Bernoulli variable with parameter $Z=s(X)$. Conditioning on a fixed $X=\boldsymbol{x}$, we have $I \sim \text{Bernoulli}(s(\boldsymbol{x}))$ and the weight $w(\boldsymbol{x})$ is a constant. Using $p_s$ as a shorthand for $Y \sim p_s(y \mid \boldsymbol{x})$, the conditional mean and variance of $W$ are 
\begin{align}
    \mathbb{E}_{p_s}[W\mid \boldsymbol{x}] &= \mathbb{E}_{p_s}[Iw(\boldsymbol{x})\mid \boldsymbol{x}] = w(\boldsymbol{x})\mathbb{E}_{p_s}[I\mid \boldsymbol{x}] = w(\boldsymbol{x})s(\boldsymbol{x}) \\
    \operatorname{Var}_{p_s}(W\mid \boldsymbol{x}) &= 
    \operatorname{Var}_{p_s}(Iw(\boldsymbol{x})\mid \boldsymbol{x}) =
    w(\boldsymbol{x})^2\operatorname{Var}_{p_s}(I\mid \boldsymbol{x}) = w(\boldsymbol{x})^2s(\boldsymbol{x})(1-s(\boldsymbol{x})).
\end{align}

Next, we can use the law of total variance\footnote{Here we again revert to slightly abused notation for compactness. In fact, we should write the total variance as $\operatorname{Var}_{X \sim p_t(\boldsymbol{x}), Y \sim p_s(y \mid \boldsymbol{x})}(W)$. However, since we consider the source distribution to be fixed, we omit $p_s$ from the left-hand side for readability.} to get
\begin{align*}
    \operatorname{Var}_{p_t}(W) &= \mathbb{E}_{p_t}\left[\operatorname{Var}_{p_s}(W\mid \boldsymbol{x})\right] + \operatorname{Var}_{p_t}\left(\mathbb{E}_{p_s}[W\mid \boldsymbol{x}]\right) \\
    &= \mathbb{E}_{p_t}[w(\boldsymbol{x})^2s(\boldsymbol{x})(1-s(\boldsymbol{x}))] + \operatorname{Var}_{p_t}(w(\boldsymbol{x})s(\boldsymbol{x})) \\
    &= \mathbb{E}_{p_t}[w(\boldsymbol{x})^2s(\boldsymbol{x})(1-s(\boldsymbol{x}))] + \mathbb{E}_{p_t}\left[(w(\boldsymbol{x})s(\boldsymbol{x}))^2\right] - \mathbb{E}_{p_t}\left[w(\boldsymbol{x})s(\boldsymbol{x})\right]^2 \\
    &= \mathbb{E}_{p_t}[w(\boldsymbol{x})^2(s(\boldsymbol{x})-s(\boldsymbol{x})^2)+ w(\boldsymbol{x})^2s(\boldsymbol{x})^2] -\mathbb{E}_{p_t}\left[w(\boldsymbol{x})s(\boldsymbol{x})\right]^2 \\
    &= \mathbb{E}_{p_t}[w(\boldsymbol{x})^2s(\boldsymbol{x}) - w(\boldsymbol{x})^2s(\boldsymbol{x})^2 + w(\boldsymbol{x})^2s(\boldsymbol{x})^2] -\mathbb{E}_{p_t}\left[w(\boldsymbol{x})s(\boldsymbol{x})\right]^2 \\
    &= \mathbb{E}_{p_t}[w(\boldsymbol{x})^2s(\boldsymbol{x})] -\mathbb{E}_{p_t}\left[w(\boldsymbol{x})s(\boldsymbol{x})\right]^2.
\end{align*}

Now, we can compare the derived variances and state that the per-observation variance of the PAPE estimator is less than or equal to that of the IW estimator if and only if 
\begin{equation}\label{eq:variance}
    \mathbb{E}_{p_t}[s^w(\boldsymbol{x})^2]-\mathbb{E}_{p_t}[s^w(\boldsymbol{x})]^2 \le \mathbb{E}_{p_t}[w(\boldsymbol{x})^2s(\boldsymbol{x})] -\mathbb{E}_{p_t}\left[w(\boldsymbol{x})s(\boldsymbol{x})\right]^2 
\end{equation}

Given that the value of the expression on the right-hand side depends heavily on the interplay between $w(\boldsymbol{x})$ and $s(\boldsymbol{x})$, and we don't know the exact relation between $s^w(\boldsymbol{x})$ and $s(\boldsymbol{x})$, it is impossible to verify whether Inequality~(\ref{eq:variance}) is true or not a priori. Here, we look at only one interesting special case, that being $p_s(\boldsymbol{x}) = p_t(\boldsymbol{x})$, which leads to a constant density ratio $w(\boldsymbol{x})=1$. In this setting, the calibrator $c$ is the identity mapping so that $f=f^w$, which means that $s(\boldsymbol{x})=s^w(\boldsymbol{x})$ pointwise. With these insights, criterion~(\ref{eq:variance}) can be written as
\begin{align}
    \mathbb{E}_{p_t}[s(\boldsymbol{x})^2]-\mathbb{E}_{p_t}[s(\boldsymbol{x})]^2 &\le \mathbb{E}_{p_t}[s(\boldsymbol{x})] -\mathbb{E}_{p_t}\left[s(\boldsymbol{x})\right]^2 \\
    \mathbb{E}_{p_t}[s(\boldsymbol{x})^2] &\le \mathbb{E}_{p_t}[s(\boldsymbol{x})].\label{eq:variance_mod}
\end{align}

Because $0 \leq s(\boldsymbol{x}) \leq 1$, it follows that 
\begin{align*}
    s(\boldsymbol{x})^2 &\le s(\boldsymbol{x}) \\ 
    \mathbb{E}_{p_t}[s(\boldsymbol{x})^2] &\le \mathbb{E}_{p_t}[s(\boldsymbol{x})], 
\end{align*}
which means that the special case criterion~(\ref{eq:variance_mod}) is always satisfied. Thus, under no shift, the per-observation variance of PAPE is always at most that of the IW estimator, making the former more sample efficient. It also gives reason to believe that this is likely the case when the shift is small, and hence the weights $w(\boldsymbol{x})$ are close to one and $s^w(\boldsymbol{x}) \approx s(\boldsymbol{x})$ pointwise. Having said that, it is possible to envision situations where the IW estimator has lower per-observation variance.  

Up to this point, we have looked only at the per-observation variances. However, it is straightforward to justify this approach as follows. We can write the variances of our estimators as variances of the means of $n$ mutually independent observations as
\begin{align*}
    \operatorname{Var}(\widehat{ACC}_{PAPE}) &= \operatorname{Var}_{p_t}\left(\frac{1}{n}\sum_{i=1}^{n}Z_i^w\right) = \frac{1}{n^2}\sum_{i=1}^{n}\operatorname{Var}_{p_t}(Z_i^w) = \frac{1}{n}\operatorname{Var}_{p_t}\left(Z^w\right) \\
    \operatorname{Var}(\widehat{ACC}_{IW}) &= \operatorname{Var}_{p_t}\left(\frac{1}{n}\sum_{i=1}^{n}W_i\right) = \frac{1}{n^2}\sum_{i=1}^{n}\operatorname{Var}_{p_t}(W_i) = \frac{1}{n}\operatorname{Var}_{p_t}(W),
\end{align*}
which means that in comparing the variances of the two estimators, it suffices to compare the per-observational variances $\operatorname{Var}_{p_t}(Z^w)$ and $\operatorname{Var}_{p_t}(W)$. 

Finally, we suggest a pragmatic empirical approximation for criterion~(\ref{eq:variance}), which can be used as a heuristic guide in choosing which estimator to use for a given (i.i.d.) sample $(X_1, X_2, ..., X_n) \sim p_t(\boldsymbol{x})^n$. This approximation replaces the expectations with empirical means, leading to  
\begin{equation*}
    \sum_{i=1}^n{s^w(X_i)^2}-\left(\sum_{i=1}^n{s^w(X_i)}\right)^2 \le \sum_{i=1}^n{w(X_i)^2s(X_i)} - \left(\sum_{i=1}^n{w(X_i)s(X_i)}\right)^2
\end{equation*}
If this inequality is true, one would favor PAPE over IW, and conversely so if it is false. Although this heuristic is not guaranteed to result in better estimates in every case, for any sufficiently large $n$, it should result in better estimates on average. If (for whatever reason) one has to choose the estimator before observing any data, a rational agent would make the (uninformative) assumption about the weights, assigning $w(\boldsymbol{x}) = 1$, which by criterion~(\ref{eq:variance_mod}) would lead them to choose PAPE. 

\section{Additional Experiments}\label{app:experiments}

\subsection{Sample Size Effect}\label{res:sample_size_effect}

The experiments described in Section \ref{sec:experiments} were run for an arbitrarily chosen evaluation data chunk size (2000 observations). In this section, we describe an ablation study, where we investigated the quality of performance estimates for different chunk sizes. We split production data into chunks of the following sizes: 100, 200, 500, 1000, 2000, and 5000. For each consecutive sample and each chunk size, the first instance of the data chunk was advanced 1,000 observations so that the first instances of each chunk for different chunk sizes were aligned. This was done to make the results between different chunk sizes comparable and to minimize the effect of changes in the underlying data. This meant that for chunk sizes of less than 1,000, not all data was used, and for chunk sizes larger than 1,000, there was some overlap between consecutive samples.

\begin{figure}
	\centering
	\includegraphics[width=0.9\textwidth]{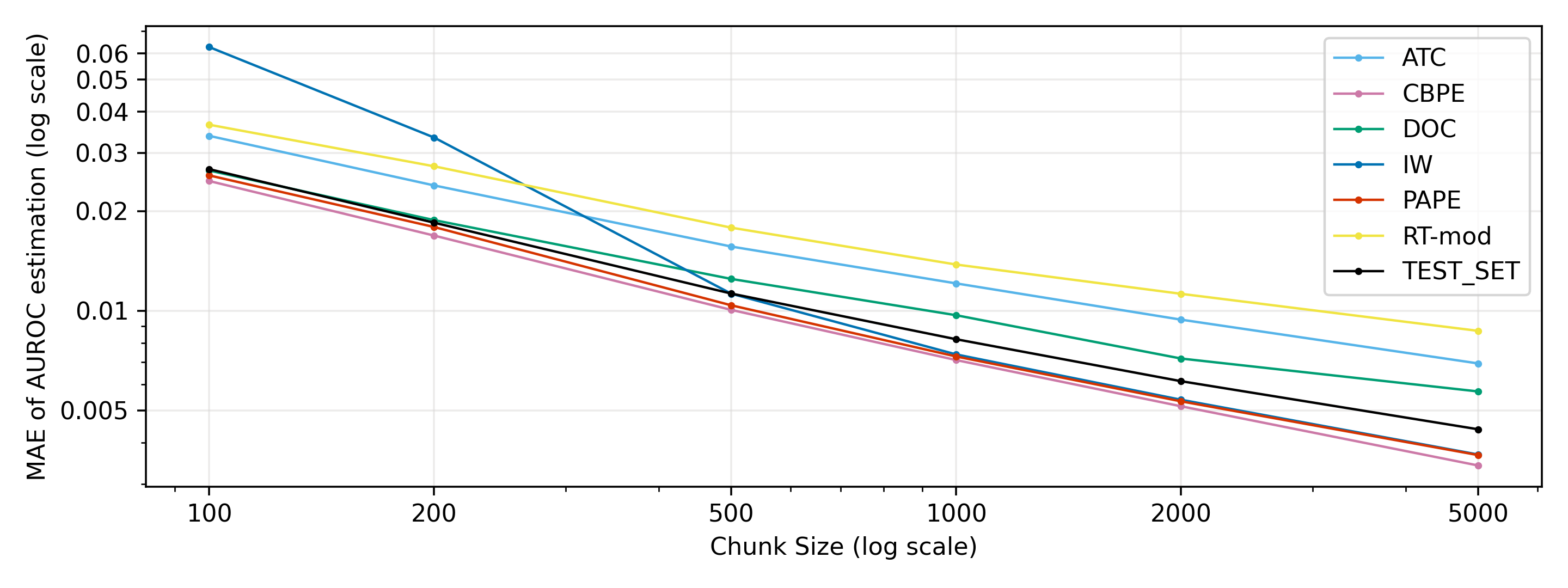}
	\caption{
		Effect of sample size on mean absolute error of AUROC estimation. Calculated for sample sizes of 100, 200, 500, 1000, 2000, and 5000, on data from California, for the prediction task: employment. 
	}
	\label{fig:chunk}
\end{figure}

Due to computational complexity, we ran the experiment only for one evaluation case. We selected the biggest data set (California), the prediction task for which performance changes significantly (ACSEmployment), the default choice algorithm for the classification task on tabular data - LGBM, and the default metric - AUROC. Since the results come from a single evaluation case, we used a typical regression metric - mean absolute error (MAE). Figure~\ref{fig:chunk} shows that all methods give less accurate AUROC estimates for smaller chunk sizes. This is expected as the random noise effects are more significant for small samples. For the evaluated case, this has the strongest impact on IW accuracy of estimation. 

\subsection{TableShift Experiments}
In addition to our main experiments with US census data, we tested PAPE and our other benchmark methods with tabular data from the recently published TableShift benchmark~\cite{gardner:2024}. Due to the defective functionality of the TableShift API, we were unable to extract all 15 datasets contained within the benchmark. Thus, we resorted to using a subset of 8 datasets we were able to retrieve, namely: ASSISTments, College Scorecard, Diabetes, Food Stamps, Hospital Readmission, Hypertension, Income, and Unemployment\footnote{These datasets are described in detail at \url{https://tableshift.org/datasets.html}}. There is some overlap between these datasets and the datasets used in Folktables~\cite{folktables}. The main differences are that the datasets in TableShift come from a wider range of providers, the datasets are generally smaller, and the datasets are preprocessed differently, in particular, to create distributional shifts between the In-Domain (ID) and Out-of-Domain (OoD) portions. These shifts do not necessarily conform to our covariate shift assumption, giving us an interesting opportunity to experiment on how PAPE performs relative to other estimators in undefined types of shift scenarios.
\begin{table}[ht!]
	\centering
	\noindent\adjustbox{max width=\textwidth}
	{
		\begin{tabular}{@{}rcccccc@{}}
			\toprule
			& \multicolumn{2}{c}{Accuracy} & \multicolumn{2}{c}{AUROC} & \multicolumn{2}{c}{F$_1$}\\
			& NMAE & NRMSE & NMAE & NRMSE & NMAE & NRMSE \\
			\midrule
			\textbf{TEST SET} & 2.11 & 2.92 & 1.03 & 1.82 & 1.01 & 1.49 \\
			\textbf{RT-mod}   & 2.22 & 3.05 & 1.39 & 2.15 & 1.19 & 1.69 \\
            \textbf{COT}    & 2.80 & 3.49 & - & - & - & - \\
			\textbf{ATC}      & 1.94 & 3.20 & 1.32 & 2.06 & 1.50 & 2.11 \\
			\textbf{DOC}      & 1.74 & 2.54 & 0.94 & 1.67 & 1.03 & 1.45 \\
			\textbf{CBPE}     & 1.79 & 2.46 & 0.97 & 1.65 & 0.77 & 1.32 \\
			\textbf{IW}       & 1.59 & 2.06 & 0.88 & 1.13 & 0.75 & 0.95 \\
			\textbf{PAPE}     & \textbf{1.52} & \textbf{1.98} & \textbf{0.83} & \textbf{1.06} & \textbf{0.67} & \textbf{0.86} \\
			\bottomrule
			\vspace{0.3pt}
		\end{tabular}
	}
	\caption{
		NMAE and NRMSE of the evaluated performance estimation methods for each estimated metric.
	}\label{tab:tableshift}
\end{table}

We used the same evaluation framework as with our main experiment in Section~\ref{sec:experiments} with the same benchmarks, implementation details, and evaluation metrics. However, the data preprocessing was handled a bit differently. We used the "training" portion of the datasets for training our models, the "validation" portion (both for ID and OoD) to train our estimators, and the "test" portion (both for ID and OoD) for evaluation. We concatenated the ID and OoD sets because for some datasets, the OoD portions were very small, sometimes less than 2,000 instances, which we used as our chunk size. The results shown in Table~\ref{tab:tableshift} align with our main experiments, showing that PAPE is superior to other estimators for all metrics tested.

\subsection{Experiments With Synthetic Data}
Since all the other experiments thus far have been performed on real-world data, where we cannot fully control the nature of the datset shift, we conducted further experiments with synthetic data created to ensure that only covariate shift is present. We start by describing the data generation process.

\subsubsection{Data Creation}
We created a spherically symmetric distribution of inputs centered at the origin of $\mathbb{R}^d$, with dimensionality $d=20$. More specifically, we generated feature vectors by first drawing Gaussian directions $G \in \mathbb{R}^{20}$ with independent $N(0, 1)$ entries and normalizing each vector to unit length, $U_i = G_i / \Vert G_i\Vert_2$, thereby ensuring isotropy on the unit sphere. Radial distances $R_i$ were then sampled independently from a half-normal distribution with $\sigma=0.15$. Each observed feature instance was then derived as $X_i = R_i \cdot U_i$, and only those with radius $R_i < 0.5$ were retained. This procedure yielded a base pool of 100,000 feature instances with uniform angular structure and a controlled radial distribution.

Next, labels for each feature instance were assigned based on a smooth, monotonic relationship between the features and the probability of the positive class. For each observed feature instance $\boldsymbol{x}$, the radius $r(\boldsymbol{x}) = \Vert\boldsymbol{x}\Vert_2$ was calculated and the conditional probability of the positive class was set to $P(y=1 \mid \boldsymbol{x}) = 1-r(\boldsymbol{x})$. The true labels $y$ were then sampled from the corresponding Bernoulli distribution. The base dataset was randomly partitioned into training (80,000) and reference (20,000) subsets, and a LightGBM classifier with default parameters was trained on the $d=20$ features using samples from the training set. 

\begin{figure}[!ht]
	\centering
	\includegraphics[width=0.64\textwidth]{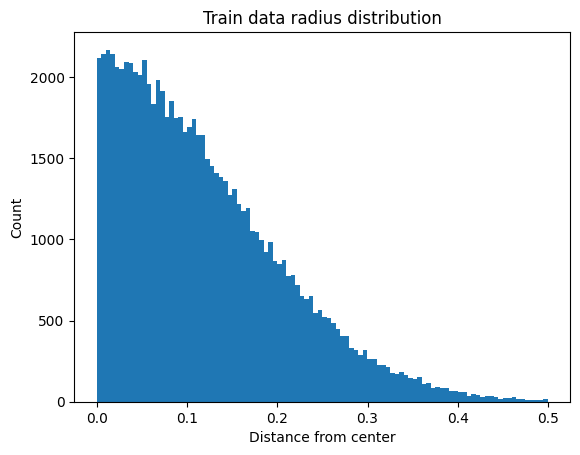}
	\caption{
		The distribution of distances from the origin in the training set for the synthetic data.
	}
	\label{fig:synthetic}
\end{figure}

The distribution of the resulting distances in the training dataset is visualized in Figure~\ref{fig:synthetic}. We see that most instances reside near the origin and are thus likely to bear the positive label. Instances farther from the origin get increasingly rarer. This results in an unbalanced dataset, where positive labels are far more common, with the fraction of positive labels in the training dataset being roughly $88\%$.

\subsection{Experiment with Gradually Increasing Covariate Shift}

In this experiment, we simulated covariate shift by repeatedly sampling production chunks of size 2,000 while gradually increasing a threshold for the radius of samples to include within the production data chunk. We let the threshold increase from 0 to 0.4 in increments of 0.025. For each resulting threshold $t$, we conducted 1,000 trials, where in each trial we sampled a chunk of data while enforcing $r(\boldsymbol{x}) > t$. Increasing the threshold is used to simulate a gradually increasing covariate shift. 
This makes predicting the right label by the monitored classifier increasingly difficult for two reasons. First, when the input data distribution shifts further away from the center, the label entropy increases since $P(y=1 \mid \boldsymbol{x})$ is equal to 1 in the center and 0.5 at the hypersphere with $r(\boldsymbol{x}) = 0.5$. Second, the regions further away from the center are less dense in the reference distribution (see Figure ~\ref{fig:synthetic}). Thus, we expect the performance of the classifier to decrease with increasing shift magnitude, and we would like our estimators to be able to catch that. 

\begin{figure}[!ht]
	\centering
	\includegraphics[width=1.0\textwidth]{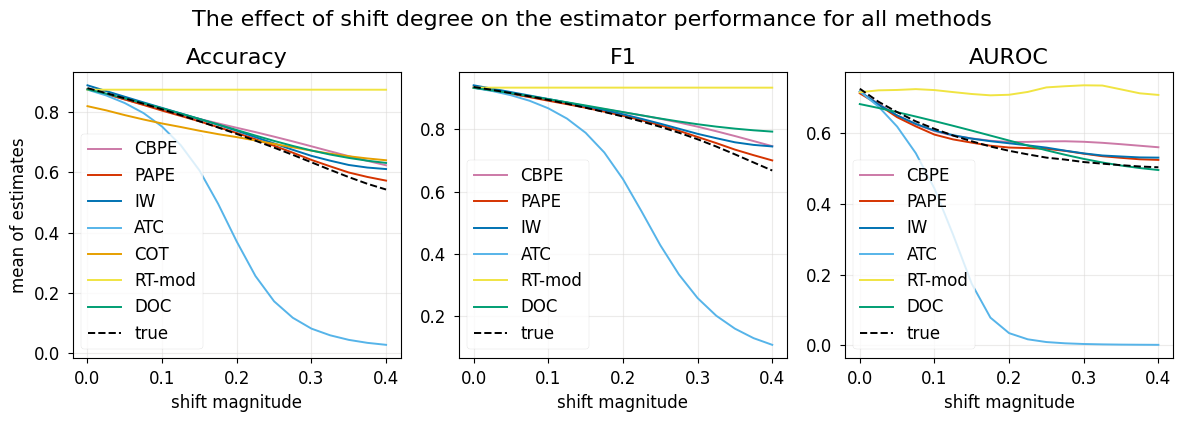}
	\caption{
		Estimator performance of each method for all three metrics with gradually increasing covariate shift. Shift magnitude corresponds to threshold $t$.
	}
	\label{fig:thresholds_all}
\end{figure}

After the 1,000 trials for each threshold, we averaged both the true model performances and the estimated performances for accuracy, F$_1$ score, and AUROC. The results are plotted in Figure~\ref{fig:thresholds_all}. 

\begin{figure}[!ht]
	\centering
	\includegraphics[width=1.0\textwidth]{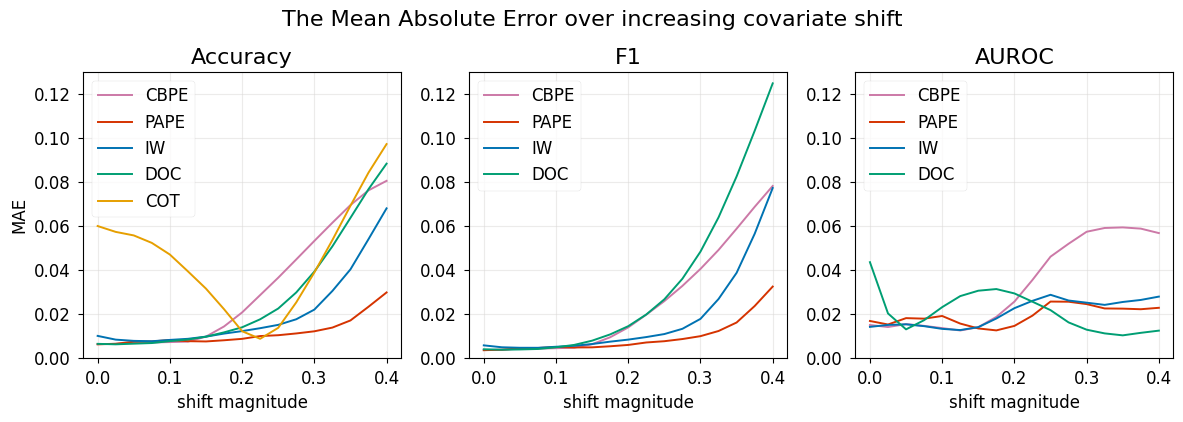}
	\caption{
		The Mean Absolute Errors of the estimators for all three metrics with gradually increasing covariate shift. 
	}
	\label{fig:thresholds_MAE}
\end{figure}

We see that both ATC and RT-mod are generating estimates that are unusable, so we decided to drop them from further analysis and focus on the rest. For the remaining methods, we plot the Mean Absolute Error (MAE) between the estimated and true metric values. The results are seen in Figure~\ref{fig:thresholds_MAE}, where we see that PAPE is consistently able to generate estimates with low error even for the most intense shifts for all three metrics and has the best overall performance of all the methods tested. 

\subsection{Experiment with Different Chunk Sizes}

\begin{figure}[!ht]
	\centering
	\includegraphics[width=1.0\textwidth]{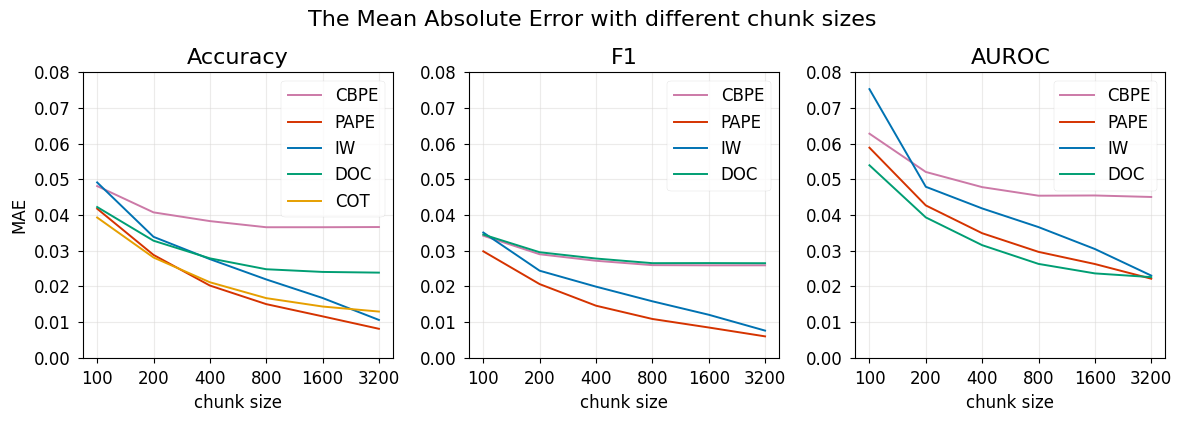}
	\caption{
		The Mean Absolute Errors for each estimator for all three metrics over different chunk sizes.
	}
	\label{fig:chunks_selected}
\end{figure}

In this experiment, we fix the threshold to 0.25 and vary the chunk size to see how it affects the performance of the estimators. We test on chunk sizes of [100, 200, 400, 800, 1600, 3200] again, performing 1,000 trials for each chunk size. As in the previous experiment with increasing covariate shift, we noted that ATC and RT-mod performed so poorly that they were left out of the comparison. For the other estimators, we show the Mean Absolute Error (MAE) between the estimated and true metric values in Figure~\ref{fig:chunks_selected}. Unsurprisingly, all methods provide better estimates with increasing chunk size, with PAPE and IW still improving while other estimators start to flat out. Overall, PAPE shows the best performance.

\end{document}